\documentclass{article}

\PassOptionsToPackage{numbers, compress}{natbib}
\usepackage[final]{nips_2017}

\usepackage[utf8]{inputenc} 
\usepackage[T1]{fontenc}    
\usepackage{hyperref}       
\usepackage{url}            
\usepackage{booktabs}       
\usepackage{amsfonts}       
\usepackage{nicefrac}       
\usepackage{microtype}      
\usepackage{times}
\usepackage{subfigure}
\usepackage{graphicx} 
\usepackage{epsfig} 

\usepackage{amsfonts,amssymb}
\usepackage{amsmath}
\usepackage{algorithm}
\usepackage{algorithmic}
\usepackage{url}
\newtheorem{theorem}{Theorem}[section]
\newtheorem{lemma}[theorem]{Lemma}

\newtheorem{definition}[theorem]{Definition}
\newtheorem{proposition}[theorem]{Proposition}

\newtheorem{corollary}[theorem]{Corollary}

\title{Convergence Analysis of Distributed Stochastic Gradient Descent with Shuffling}

\author{
Qi Meng$^{1,}$\thanks{This work was done when the first and third author were visiting Microsoft Research Asia.}, Wei Chen$^2$, Yue Wang$^{3,*}$, {Zhi-Ming Ma$^4$, Tie-Yan Liu$^2$}\\
$^1$Peking University\quad$^2$Microsoft Research\quad$^3$Beijing Jiaotong University\\$^4$Chinese Academy of Mathematics and Systems Science	\\
$^1$qimeng13@pku.edu.cn;
$^2$\{wche, tie-yan.liu\}@microsoft.com;\\
$^3$11271012@bjtu.edu.cn; $^4$mazm@amt.ac.cn
}

\begin{document}
	
	\maketitle
	
	\begin{abstract}
		When using stochastic gradient descent (SGD) to solve large-scale machine learning problems, a common practice of data processing is to shuffle the training data, partition the data across multiple threads/machines if needed, and then perform several epochs of training on the re-shuffled (either locally or globally) data. The above procedure makes the instances used to compute the gradients no longer independently sampled from the training data set, which contradicts with the basic assumptions of conventional convergence analysis of SGD. Then does the distributed SGD method have desirable convergence properties in this practical situation? In this paper, we give answers to this question. First, we give a mathematical formulation for the practical data processing procedure in distributed machine learning, which we call \emph{(data partition with) global/local shuffling}. We observe that global shuffling is equivalent to without-replacement sampling if the shuffling operations are independent. 
		We prove that SGD with global shuffling has convergence guarantee in both convex and non-convex cases. An interesting finding is that, the non-convex tasks like deep learning are more suitable to apply shuffling comparing to the convex tasks. Second, we conduct the convergence analysis for SGD with local shuffling. The convergence rate for local shuffling is slower than that for global shuffling, since it will lose some information if there's no communication between partitioned data. Finally, we consider the situation when the permutation after shuffling is not uniformly distributed (We call it insufficient shuffling), and discuss the condition under which this insufficiency will not influence the convergence rate. Our theoretical results provide important insights to large-scale machine learning, especially in the selection of data processing methods in order to achieve faster convergence and good speedup. Our theoretical findings are verified by extensive experiments on logistic regression and deep neural networks.
	\end{abstract}
	
	\section{Introduction}
	
	Stochastic gradient descent (SGD) is a widely-used optimization technology in many applications, especially in deep learning, due to its simplicity and good empirical performances \cite{rakhlin2011making,bottou2010large,dean2012large}. The goal of SGD is to minimize the empirical risk, which is defined as the averaged loss of a model over $n$ training samples. By exploiting the additive nature of the empirical risk function, SGD estimates the full gradient of the model over all the training samples by using only a subset or a random sample of the training data and updates the model towards the negative direction of the estimated gradient (stochastic gradient). In large-scale applications (i.e., $n$ is large), SGD benefits from lower computational complexity at each iteration although it may require more iterations to converge than the full gradient descent method (GD), mainly due to the non-negligible influence (i.e., variance) introduced by the stochastic strategy. 
	
	In the literature, the theoretical properties of SGD, especially its convergence rate, has been well studied. A common assumption used in these studies is the (i.i.d.) sampling of the training data when estimating the full gradient, which can ensure that the stochastic gradient is an unbiased estimator of the full gradient \cite{rakhlin2011making,bottou2010large, ghadimi2013stochastic}. Recently, several literatures work on the convergence properties of SGD when the training data is shuffled or without-replacement sampled in sequential setting. However, to the best of our knowledge, there is no previous analysis for SGD with the practical data processing in the parallel setting.
	
	In practice, with the emergence of large-scale data and complex models like deep neural networks, one single machine becomes insufficient to store all the training data and the sequential training process becomes unacceptably slow. As a consequence, in recent years, distributed machine learning has become a trend. A widely-used data processing method for distributed machine learning is as follows: (1) shuffle the training data, (2) evenly partition the data into non-overlapping subsets and allocate each subset to a local worker (either a thread or a machine), (3) do SGD training (with necessary communications between workers to ensure global consensus of the learned model), (4) reshuffle the dataset after each training epoch, either locally or globally. \footnote{If the training data can be centrally stored, global shuffling is widely used, where the entire training data set is shuffled after each epoch. Otherwise, local shuffling is preferred to save communication cost, where each local worker randomly shuffles its own subset of training data after each epoch. \cite{lvsr,nyu-dl,GroundHog}}. If it is done globally, repeat (2) and (3). If it is done locally, repeat (3). For ease of discussion, we refer to the above data processing procedure as \emph{(data partition with) shuffling}, and that with global shuffling or local shuffling after each epoch as \emph{global shuffling} or \emph{local shuffling}, respectively. 
	
	Compared with i.i.d sampling, data partition with shuffling forces the algorithm to process more different training instances. Intuitively, this leads to a model with better empirical performance \cite{gurbuzbalaban2015random}\cite{recht2012toward}. Shuffling algorithms aim to produce a random permutation. A widely used shuffling algorithm is Fisher-Yates shuffle, whose asymptotic complexity is $O(n)$. It is equivalent to without-replacement sampling. Fisher-Yates shuffle may introduce potential bias and implementation errors to make the permutation are not uniformly distributed in practice \cite{fisher1938statistical}. There are also other shuffling algorithms like Top-to-Random shuffle and Riffle shuffle \cite{levin2009markov}. In this paper, we refer sufficient shuffling (or shuffling) to a shuffling algorithm which produces a permutation distributed from uniform distribution.  Sufficient shuffling is an ideal case. In practice, shuffling algorithms may produce potential bias \cite{fisher1938statistical}. If the shuffling is insufficient, how it effects the training procedure is a problem. If it won't effect too much, we can neglect the potential bias directly or design more efficient shuffling algorithms for large-scale learning tasks with some bias.      
	
	In this paper, we aim to provide the convergence guarantee for distributed SGD with the practical data partition with shuffling, discuss its speedup condition, and compare it with i.i.d sampling. First, we give the explicit mathematical description of data partition with global/local shuffling. We observed that, without-replacement sampling is one of the methods to produce a random permutation and all the data partition with sufficient global shuffling in the parallel setting is equivalent to without-replacement sampling if the shuffling operations are independent. Based on that, we prove the convergence rates of distributed SGD in both convex and non-convex cases with \emph{global shuffling}. 
	We have the following conclusions for global shuffling. (1) For convex and strongly-convex cases, if the number of epochs is relatively small with respect to the difficulty of optimizing the objective function (e.g. the condition number is large in the strongly-convex case), global shuffling is comparable to i.i.d sampling. If the training goes on several epochs, global shuffling has speedup guarantee. \footnote{We will explain the reason for the nonoverlap of the two conditions later.} (2) In the non-convex case, under the mild condition that the number of training epochs is smaller than the size of training data, global shuffling is comparable with sampling, and at the same time a linear speedup can be achieved. These conclusions suggest that, shuffling is especially suitable for the difficult non-convex NN models in terms of convergence rate. 
	
	Second, based on our convergence analysis for global shuffling, we prove a convergence rate for \emph{local shuffling} in both convex and non-convex cases. For the non-convex case, the convergence rate of local shuffling is comparable with that of global shuffling; for the easier convex case, it is worse than global shuffling. The reason is, comparing to global shuffling, since it will lose some information if there's no communication between partitioned data. These disadvantages are more significant for the easier convex optimization tasks. 
	
	Finally, we prove a convergence rate of distributed SGD with \emph{insufficient global/local shuffling}. In addition, we will get a shuffling result by much smaller amounts of complexity, which is although not sufficient, but very close to the random shuffling. We prove that, if the insufficient error of the shuffling can be upper bounded by {\small$\sqrt{bM}/n$} (for global shuffling) or {\small$M\sqrt{b}/n$} (for local shuffling), where $n$ is the size of the training data, $b$ is the local mini-batch size and $M$ is the number of local computation nodes, the insufficiency of the shuffling will not influence the convergence rate.
	
	We conduct experiments on logistic regression and neural networks. The empirical results are consistent with our theoretical findings, which indicates that our theoretical analysis lays down a good foundation for SGD training in practice.
	
	\section{Distributed SGD with shuffling and related work}

	\subsection{ Distributed SGD with shuffling}
	In this section, we will briefly introduce distributed SGD, describe data partition with random shuffling, and discuss its relationship with sampling strategy. 
	Suppose that we have a training data set {\small$S=\{(x_1,y_1),...,(x_n,y_n)\}$} with $n$ instances i.i.d. sampled from $\mathcal{Z}=\mathcal{X}^d\times\mathcal{Y}$ according to the underlying distribution $\mathcal{P}$. The goal is to learn a good prediction model $h(w)\in\mathcal{F}: \mathcal{X}^d\to\mathcal{Y}$ which is parameterized by $w$. The prediction accuracy for instance $(x,y)$ is measured by a loss function $f(y,h(w,x))$. 
	Stochastic Gradient Descent (SGD) \cite{rakhlin2011making,bottou2010large,zinkevich2010parallelized} is a very popular optimization algorithm, which aims to minimize the empirical risk {\small$
		F(w)=\frac{1}{n}\sum_{i=1}^{n}f(y_i,h(w,x_i))
		$} on the training data.
	For simplicity, we denote $f(y_i,h(w,x_i))$ as $f_i(w)$. In each iteration, SGD updates the model parameter $w$ towards the negative direction of the stochastic gradient, i.e.,
	{\small$
		w_{t+1}=w_t-\eta_t\nabla f_{i_t}(w_t),$}	where $\eta_t$ is the learning rate and $i_t$ is a randomly sampled training instance. 
	
	The most commonly used sampling methods are i.i.d. sampling \cite{rakhlin2011making}\cite{dekel2012optimal} and without-replacement sampling \cite{shamir2016without}\cite{bottou2009curiously}. The mathematical description for i.i.d. sampling is {\small$P(i_t=j)=1/n, \forall j\in[n]$} and for \emph{without-replacement sampling} is {\small$P(i_t=j)=1/(n-t+1), \forall j\in [n]/\{i_1,\cdots,i_{t-1}\}.$}
	In the literature of convergence analysis on SGD, most works assume i.i.d. sampling \cite{rakhlin2011making,ghadimi2013stochastic,bousquet2008tradeoffs, li2014efficient,dekel2012optimal}. Only very recently, a convergence rate of SGD with without-replacement sampling is analyzed \cite{shamir2016without} by using transdunctive Rademacher complexity, in the case that the number of epochs is equal to one.
	
	In large-scale machine learning, if the number of data is large and the full data cannot be stored by one local worker, the training process is very time-consuming. To utilize multiple computational nodes to speed up the training process, distributed machine learning algorithms are designed \cite{zinkevich2010parallelized}\cite{li2014efficient}\cite{dean2012large}\cite{recht2013parallel} and distributed SGD is widely-used for the optimization. Distributed SGD could be implemented in either a synchronous or an asynchronous manner. In this paper, we will focus on the synchronous implementation due to its popularity and promising performance in practice \cite{li2014efficient,dekel2012optimal}. \footnote{Our results can be extended to the asynchronous implementation, with some extensive efforts. We leave that for the future work.}
	To be specific, at iteration $t$, local worker $m$ computes the gradient over a local subset of instances $b_t^m$ with size $b$ and all the local gradients are averaged to update the model parameter, i.e., 
	{\small\begin{equation}\label{SSGD}
		w_{t+1}=w_t-\eta_t\cdot\frac{1}{Mb}\sum_{m=1}^M \sum_{i\in b_t^m}\nabla f_{i}(w_t).
		\end{equation}}However, we would like to point out that the local subset of instances $b_t^m$ in practical distributed machine learning is neither i.i.d. sampled nor in a without-replacement manner. A widely-used strategy for distributed machine learning is as follows: 
	
	(1) Shuffle the training data $[n]$ into $\sigma([n])$, where $\sigma$ is a random permutation operator. 
	
	(2) Evenly partition the shuffled training data $\sigma([n])$ into $M$ parts \emph{in order}, i.e., {\small$D_m\triangleq\{\sigma([n])_{(m-1)n/M+1}, \cdots, \sigma([n])_{(m)n/M }\},$} where $m\in[M].$ Allocate the subset $D_m$ to local worker $m$.
	
	(3) Do SGD training, with the instances sequentially processed in $D_m$. That is, $b_t^m=D_m(t):=\{\sigma([n])_{(m-1)n/M+(t-1)b+1},\cdots,\sigma([n])_{(m-1)n/M+tb}\}$. 
	
	(4) Reshuffle the data globally when all the data have been processed but the training curve does not converge yet and go back to step (1), or reshuffle the data locally and go back to step (3).
	Specifically, we perform a random permutation on the local data $D_m$ into $\sigma_m(D_m)$ and continue the SGD training. 
	
	For ease of presentation, we call the above data allocation strategy in distributed SGD \emph{data partition with shuffling} and refer to its two versions in step (4) as \emph{global shuffling} and \emph{local shuffling} respectively. Furthermore, we denote the data steam in the {\small$S$} epochs on {\small$M$} local workers as {\small$RS_g([n];S,M)$} and {\small$RS_l([n];S,M)$} respectively. More details could be found in Algorithm 1.
	Random shuffling aims to let each permutation of $[n]$ have equal probability, which means that the distribution of the random variable $\sigma([n])$ is uniform among all the permutations of $[n]$. If it is the case, we call it \emph{sufficient shuffling}. In the rest of the paper, we regard the shuffling as sufficient shuffling if it does not point out that the shuffling is insufficient. 
	We observe that data partition with sufficient random shuffling is equivalent to without-replacement sampling if the shuffling operations are independent (as shown by Proposition 1.1 in Appendix), which is helpful for us to prove the convergence rate. 	
	\begin{algorithm}
		\caption{Distributed SGD with Random Shuffling}
		\label{Alg1}
		\begin{algorithmic}
			\REQUIRE initial vector $w_0^1$, size of local mini-batch $b$, number of iterations $T=n/bM$ for each epoch, number of epochs $S$, number of local workers $M$.
			\ENSURE $w^S$ or {\small$\bar{w}^S=\frac{1}{TS}\sum_{s=1}^S\sum_{t=1}^Tw_t^s$}
			\FOR{$s=1,2,...,S$}
			\STATE {\small$w_0^{s+1}=w_T^s$}.
			\STATE \textbf{Option 1:} \emph{For master:} Randomly shuffle the full data and partition them into $M$ non-overlapped subsets, then allocate the $m$-th subset to the $m$-th local worker.
			\STATE \textbf{Option 2:} \emph{For local work $m$:} Randomly shuffle the local data $D_m$.
			\FOR{$t=0,1,...,T$}
			\STATE \emph{For local worker $m$:}
			\STATE \textbf{Pull} {$w_t^s$ from the master.}
			\STATE \textbf{Compute} {{\small$\nabla f_{D_m^s(t)}(w_t^s)=\sum_{i\in D_m^s(t)}\nabla f_i(w_t^s)$} and \textbf{push} it to the master.}
			\STATE \emph{For master:}
			\STATE \textbf{Update} {\small$w_{t+1}^s=w_t^s-\eta_t^s\cdot\frac{1}{bM}\sum_{m=1}^M\nabla f_{D_m(t)^s}(w_t^s).$}
			\ENDFOR
			\ENDFOR
		\end{algorithmic}
	\end{algorithm} 
	\subsection{Related Work}
	Shamir has studied SGD with without-replacement sampling with one epoch in sequential setting by using transductive Rademacher Complexity in \cite{shamir2016without}. Our target is distributed SGD with global shuffling (equivalent to without replacement sampling with multiple epochs), local shuffling, insufficient shuffling in parallel setting. In addition, we have also studied non-convex case which is not studied in \cite{shamir2016without}. \cite{recht2012toward} investigated SGD with without-replacement sampling for least means square optimization and showed that the convergence rate of SGD with without-replacement sampling is faster than $i.i.d$ sampling by using arithmetic-geometric mean inequality. However, the results are only suitable for least mean squares optimization. \cite{gurbuzbalaban2015random} has studied SGD with random shuffling. The convergence rate depends on the times of reshuffling, but there's no explicit description on how the size of training data or the number of iterations influence the convergence rate. Our results show that the convergence rate of distributed SGD with shuffling is related to the number of epochs (the times of reshuffling), the number of training data, the number of local machines and the number of mini-batch size. Moreover, we also study the convergence rates for distributed SGD with local shuffling methods and insufficient shuffling methods, which are not contained in the related works.

	\section{Convergence analysis of distributed SGD with global shuffling}
	In this section, we will analyze the convergence rate of distributed SGD with global shuffling, for both convex and non-convex cases. 
	In order to perform the convergence analysis, we take the following commonly used definitions and assumption in optimization \cite{rakhlin2011making}\cite{johnson2013accelerating}\cite{ghadimi2013stochastic}\cite{nesterov2013introductory}.
	\begin{definition} $F(w)$ is $L$-Lipschitz about $w$ if
		{\small$
			|F(w_1)-F(w_2)|\leq L||w_1-w_2||,  \forall w_1,w_2.$}
		$F(w)$ is $\rho$-smooth about $w$ if
		{\small$
			F(w_1)\leq F(w_2)+\langle\nabla F(w_2),w_1-w_2\rangle+\frac{\rho}{2}\|w_1-w_2\|^2.
			$}
	\end{definition}	
	\begin{definition} $F(w)$ is convex about $w$ if
		{\small$
			F(w_1)\geq F(w_2)+\langle\nabla F(w_2),w_1-w_2\rangle.$}
		$F(w)$ is $\mu$-strongly convex about $w$ with positive coefficient $\mu$ if
		{\small$
			F(w_1)\geq F(w_2)+\langle\nabla F(w_2),w_1-w_2\rangle+\frac{\mu}{2}\|w_1-w_2\|^2.$}
	\end{definition}
	\textbf{Assumption 1:} {\small$\|\nabla F(w)\|^2\leq G^2$} and {\small$\|\nabla f_i(w)\|^2\leq B^2,\forall w, i$}.
	
	We use $\kappa=\rho/\mu$ to denote the condition number. In our analysis, we use transductive Rademacher Complexity \cite{el2009transductive} as a tool (whose definition is shown in Definition 3.1 in the Appendix) to show the convergence rate for distributed SGD with global shuffling in both convex and non-convex cases. The detailed proofs of convex and nonconvex cases are shown in the Appendix due to space limitation.
	\subsection{The strongly-convex case}
	The following theorem shows the convergence rate of distributed SGD with global shuffling in the strongly convex case.
	\begin{theorem}\label{thm2}
		Suppose the objective function is strongly convex and smooth, and Assumption 1 holds. Then distributed SGD with global shuffling {\small$RS_g([n],S,M)$} 
		and learning rate {\small$\eta_t^s= \frac{2}{\mu((s-1)T+t)}$} where {\small$T=\frac{n}{bM}$}, has the following convergence rate :
		{\small\begin{align*}
			&\mathbb{E}\|w^{S}-w^*\|^2\leq\mathcal{O}\left(\min\left\{\frac{bM}{Sn},\frac{\kappa^2(bM)^2\log{Sn}}{(Sn)^2}+\frac{\kappa^2bM\log{n}}{Sn^2}\right\}+\frac{\log{n}}{n}\right).
			\end{align*}}
	\end{theorem}
	The main idea of the proof is as follows. Due to the non-i.i.d property brought by shuffling, there is bias and the variance becomes large. The variance can be further decomposed to the bias between shuffling and i.i.d sampling, which is upper bounded by using transductive Rademacher Complexity, and the variance of a subset of data, which can be bounded by using Theorem B in page 208 \cite{rice2006mathematical}.

	Based on the above theorem, we analyze the conditions for: (1) achieving the same convergence rate with distributed SGD with i.i.d sampling; (2) the speedup compared with sequential SGD with shuffling.\footnote{The results ignore the $\log$ term.} To be specific, the convergence rate for distributed SGD with i.i.d sampling is {\small$\mathcal{O}\left(\min\{\frac{bM}{Sn},\frac{b^2M^2\kappa^2\log{n}}{(Sn)^2}\}+\frac{1}{Sn}\right)$} (Shown in Theorem 7.1 in Appendix). If the additional terms {\small$\frac{\log{n}}{n}$} and {\small$\frac{\kappa^2bM\log{n}}{Sn^2}$} in the bound for distributed SGD does not dominate the bound, then the convergence rate for global shuffling are comparable with that for sampling. Assume that sequential SGD needs $S$ epochs to achieve a target training accuracy. If distributed SGD needs {\small$\alpha S$} epochs to achieve the same accuracy as sequential SGD, its speedup ratio is {\small$M/\alpha$} because the calculation time for distributed SGD is about $M$ times faster than sequential SGD. If the term {\small$\frac{\log{n}}{n}$} dominate the bound, both distributed SGD and sequential SGD has dominated term {\small$\frac{\log{n}}{n}$}. Thus it will achieve linear speedup.  Thus we have the following corollary.
	\begin{corollary}
		If {\small$S\leq \frac{bM\kappa^2}{n}$}, the convergence rate of distributed SGD with global shuffling {\small$RS_g([n],S,M)$} is comparable with distributed SGD with i.i.d sampling.
		If {\small$S\geq bM\max\{1,\frac{\kappa^2}{n}\}$}, distributed SGD with global shuffling {\small$RS_g([n],S,M)$} achieves at least linear speedup compared with the sequential SGD with {\small$RS_g([n],S,1)$}.
	\end{corollary}
	
	\subsection{The convex case}\label{sec3.1}
	The following theorem shows the convergence rate of distributed SGD with global shuffling in the convex case, which extends Corollary 1 in \cite{shamir2016without} to the mini-batch and multi-epoch case.
	\begin{theorem}\label{convex thm}
		Suppose the objective function is convex and $L$-Lipschitz, and Assumption 1 holds. Then distributed SGD with global shuffling {\small$RS_g([n],S,M)$} and learning rate {\small$\eta_t^s=\sqrt{\frac{L}{((s-1)T+t)}}$} where $T=\frac{n}{bM}$, has the following convergence rate:
		{\small$
			\mathbb{E}F(\bar{w}^S)-F(w^*)\leq\mathcal{O}\left\{\frac{1}{\sqrt{nS}}+\frac{Mb}{nS}+\sqrt{\frac{1}{n}}\right\}.
			$}
	\end{theorem}
	Please note that the convergence rate of distributed SGD with i.i.d sampling is {\small$\mathcal{O}\left(\sqrt{\frac{1}{nS}}+\frac{bM}{nS}\right)$} \cite{li2014efficient} for $nS$ effective passed data. With the similar analysis in strongly-convex case, we have the following corollary. 
	
	\begin{corollary}
		If {\small$S\leq \frac{Mb}{\sqrt{n}}$}, the convergence rate of distributed SGD with global shuffling {\small$RS_g([n],S,M)$} is comparable with distributed SGD with i.i.d sampling.
		If {\small$S>\frac{Mb}{\sqrt{n}}$}, distributed SGD with global shuffling {\small$RS_g([n],S,M)$} achieves linear speedup compared with the sequential SGD with {\small$RS_g([n],S,1)$}.
	\end{corollary}

	\subsection{Non-convex case}\label{section3.2}
	The following theorem characterizes the convergence rate of distributed SGD with global shuffling for non-convex objective functions.
	\begin{theorem}\label{thm3}
		Suppose the objective function is non-convex and $\rho$-smooth, and Assumption 1 holds. By setting {\small$\eta=\min\left\{\frac{1}{\sqrt{ST}}\cdot\sqrt{\frac{2(F(w_0^1)-F(w^*))}{\frac{3\rho B^2}{bM}\left(1+\frac{584\log{T}}{T}\right)}},\frac{1}{6\rho}\right\}$} where $T=\frac{n}{bM}$, distributed SGD with global shuffling {\small$RS_g([n],S,M)$} has the following convergence rate:
		{\small$
			\frac{1}{TS}\left(\sum_{s=1}^S\sum_{t=1}^T\|\nabla F(w_t^s)\|^2\right)\leq\mathcal{O}\left(\sqrt{\frac{(F(w_0^1)-F(w^*))\rho}{Sn}}+\frac{\log{(n)}}{n}\right).
			$}
	\end{theorem}
	Considering the convergence rate for distributed SGD with i.i.d sampling in non-convex cases is {\small$\mathcal{O}\left(\sqrt{1/Sn}\right)$} \cite{ghadimi2013stochastic}, we have the following corollary based on the above theorem.
	\begin{corollary}
		If {\small$S<n$}, the convergence rate of distributed SGD with global shuffling {\small$RS_g([n],S,M)$} is comparable with distributed SGD with i.i.d sampling.
		If {\small$S<n$}, distributed SGD with global shuffling {\small$RS_g([n],S,M)$} achieves linear speedup compared with the sequential SGD with {\small$RS_g([n],S,1)$}.
	\end{corollary}

	\subsection{Conclusions}
	
	We have the following conclusions based on all the results in convex and non-convex cases.
	(1) For convex and strongly-convex cases, if the training is insufficient with respect to the difficulty of optimizing the objective function (e.g. the number of epochs is not very large in the convex case and the condition number is large in the strongly-convex case), global shuffling is comparable to sampling. If the training is sufficient relatively, global shuffling has speedup guarantee. It is interesting that the condition to achieve speedup contradict the condition to achieve the comparable performance with i.i.d sampling for both convex and strongly convex cases. It is because that the speedup condition for distributed SGD in convex cases is strong, either with shuffling or sampling. The intuitive explanation is that the convergence rate is fast in convex cases and its speedup is limited.  
	
	(2) In the non-convex case, under the mild condition that the number of epochs is smaller than the number of training data, global shuffling is comparable with i.i.d. sampling, and at the same time a linear speedup can be achieved. These conclusions suggest that, shuffling, which is much more efficient for large data in deep learning in terms of computational complexity, is also especially suitable for the difficult non-convex NN models in terms of convergence rate.

	\section{Convergence rate of distributed SGD with local shuffling}\label{local} 
	Compared to global shuffling, local shuffling method will lose some information because there's no data communication between local workers. It is efficient to implement and saves communication time. However, the convergence rate will be slowed down. The negative impact brought by local shuffling is reflected by the term which measures the unbiasedness caused by the non-i.i.d property of shuffling. This term should be bounded by transductive Rademacher Complexity. If we use local shuffling instead of global shuffling, the upper bound of the term is determined by the local training size $n/M$ instead of the global training size $n$. Intuitively, the information gain brought by the data decreases because there's no data communication between local workers. The following theorem gives the convergence rates of distributed SGD with local shuffling in the convex and non-convex cases. 
	\begin{theorem}
		Conditioned on the partition, the expected convergence rate of distributed SGD with local shuffling in the convex and Lipschitz case is similar to the results given in Theorem \ref{convex thm}, with the term $\sqrt{\frac{1}{n}}$ replaced by $\sqrt{\frac{M}{n}}$; in the strongly convex and smooth case is similar to the results given in Theorem \ref{thm2}, with the term $\frac{\log {n}}{n}$ replaced by $\frac{M\log{n}}{n}$; in the non-convex and smooth case is similar to the results given in Theorem \ref{thm3}, with the term $\frac{\log{n}}{n}$  replaced by $\frac{M\log{n}}{n}$.
	\end{theorem} 
	
	\textbf{Discussion:} From the above theorem, we can see that by using local shuffling, the convergence rate for the convex cases will be dominated by the local training size. Enlarging the number of machines and the number of epochs will not help to improve the convergence rate. For the non-convex case, the convergence rate is relatively slower, thus the negative influence brought by local shuffling will not dominate the convergence rate. However, the speedup condition becomes $S<\frac{n}{M}$, which is $M$ times worse than that for global shuffling. Saying that, the condition is still easy to be satisfied in most practical situations. 
	
	\section{ Convergence rate of distributed SGD with insufficient shuffling} 
	As we know that sufficient shuffling is an ideal case. In general, the shuffling is insufficient. We introduce shuffling error $\epsilon(A,n)$, where $A$ is the shuffling algorithm to measure the gap between sufficient and insufficient shuffling. The shuffling error is defined as the total variation distance between output distribution $v_{\pi([n])}(A,n)$ from shuffling algorithm $A$ and the uniform distribution $u_{\pi([n])}$, where $\pi([n])$ is the permutation set contained all the permutation on the $n$ training instances. We put its explicit formulation in the Appendix (Definition 2.1).
	
	The following proposition establishes the gap between the conditional distribution of $t$-th minibatch under the uniform distribution and the distribution after insufficient shuffling. The difference for the conditional probability can be upper bounded by the shuffling error $\epsilon(A,n)$. 
	
	\begin{proposition}\label{propo311}
		Denote {\small$\mathbb{P}_{\cdot|t}(B_m,m\in[M])$} as the conditional probability {\small$\mathbb{P}_\cdot(D_m(t+1)=B_m,m\in[M]|D_m(1),\cdots,D_m(t)\neq B_1,\cdots,B_M, m\in[M])$}. If the shuffling error {\small$\epsilon(A,n)\leq\frac{bM}{n}$}, then for $t+1<T$, we have
		{\small$
			|\mathbb{P}_{v|t}(B_m,m\in[M])-\mathbb{P}_{u|t}(B_m,m\in[M])|\leq\frac{4n\epsilon(A,n)}{n-bMt}.
			$}
	\end{proposition} 
	We consider the situation when the shuffling method is not sufficient by using Proposition \ref{propo311} as a bridge. The following theorem gives the convergence rates for insufficient global shuffling in both the convex and non-convex cases. The results for insufficient local shuffling are just similar, and we omit it due to space restrictions.
	\begin{theorem}
		Suppose Assumption 1 holds. Distributed SGD with insufficient global shuffling $R_g([n];S,M)$ and learning rate in convex and $L$-Lipschitz continuous case has the following convergence rate,
		{\small$
			\mathbb{E}F(\bar{w}^S)-F(w^*)\leq\mathcal{O}\left\{\frac{1}{\sqrt{Sn}}+\frac{bM}{Sn}+\sqrt{\frac{1}{n}}+\epsilon(A,n)\ln{n}\right\}.
			$}	Set the learning rate $\eta_t^s=\frac{2}{\mu((s-1)T+t)}$ in strongly convex and smooth case, it has the following convergence rate,
		{\small$
			\mathbb{E}\|w^{S}-w^*\|^2\leq\mathcal{O}\left(\frac{\log{n}}{n}+\frac{n\epsilon(A,n)^2}{bM}\right)+\mathcal{O}\left(\min\left\{\frac{bM}{Sn},\frac{\kappa^2(bM)^2\log{Sn}}{(Sn)^2}+\frac{\kappa^2bM\log{n}}{Sn^2}+\frac{\kappa^2n\epsilon(A,n)^2}{SbM}\right\}\right).
			$}
		
		Set {\small$\eta=\min\left\{\frac{1}{\sqrt{ST}}\sqrt{\frac{2(F(w_0^1)-F(w^*))}{\frac{3\rho B^2}{bM}\left(1+\frac{584\log{T}}{T}\right)}},\frac{1}{6\rho}\right\}$} in non-convex case,
		it has the following convergence rate
		{\small$
			\frac{1}{TS}\left(\sum_{s=1}^S\sum_{t=1}^T\|\nabla F(w_t^s)\|^2\right)
			\leq\mathcal{O}\left(\sqrt{\frac{1}{Sn}}+\frac{\log{n}}{n}+\frac{n(\epsilon(A,n))^2}{bM}\right).
			$}
	\end{theorem}
	
	\textbf{Discussion:} From the above theorem, we can get that if {\small$\epsilon(A,n)\leq\frac{\sqrt{bM}}{n}$}, the terms related to shuffling error {\small$\epsilon(A,n)$} will not dominate the bounds, and therefore insufficiency of the global shuffling algorithm will not influence the convergence rate. Similarly, we can get that if {\small$\epsilon(A,\frac{n}{M})\leq\frac{M\sqrt{b}}{n}$}, the insufficiency of the local shuffling will not influence the convergence rate.

	\section{Experiments}
	\begin{figure*}[t!]
		\centering
		\subfigure[]{
			\label{fig1a}
			\includegraphics[width=1.3in]{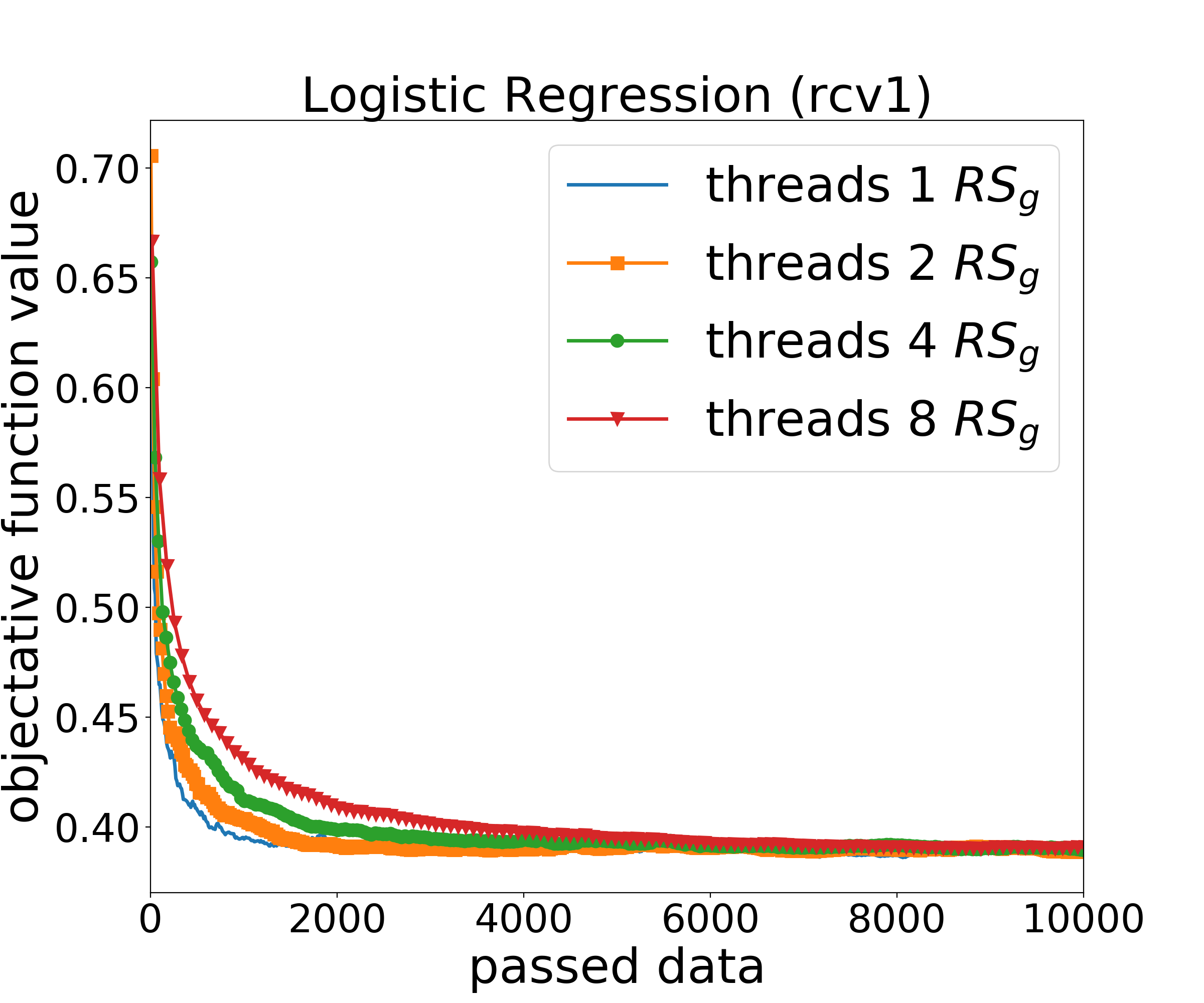}}
		\subfigure[]{
			\label{fig1c}
			\includegraphics[width=1.3in]{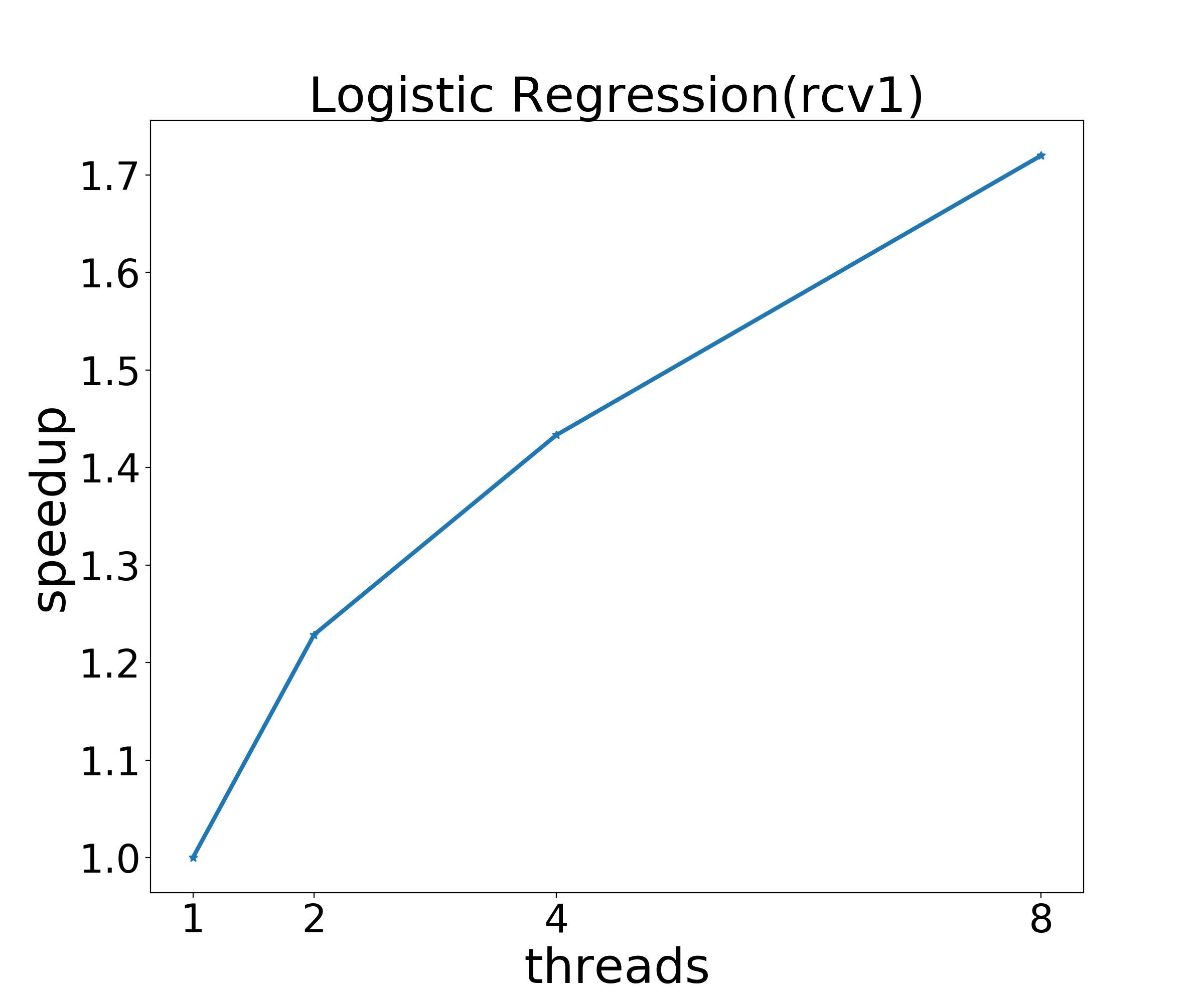}}
		\label{Fig1}
		\subfigure[]{
			\label{fig2a}
			\includegraphics[width=1.3in]{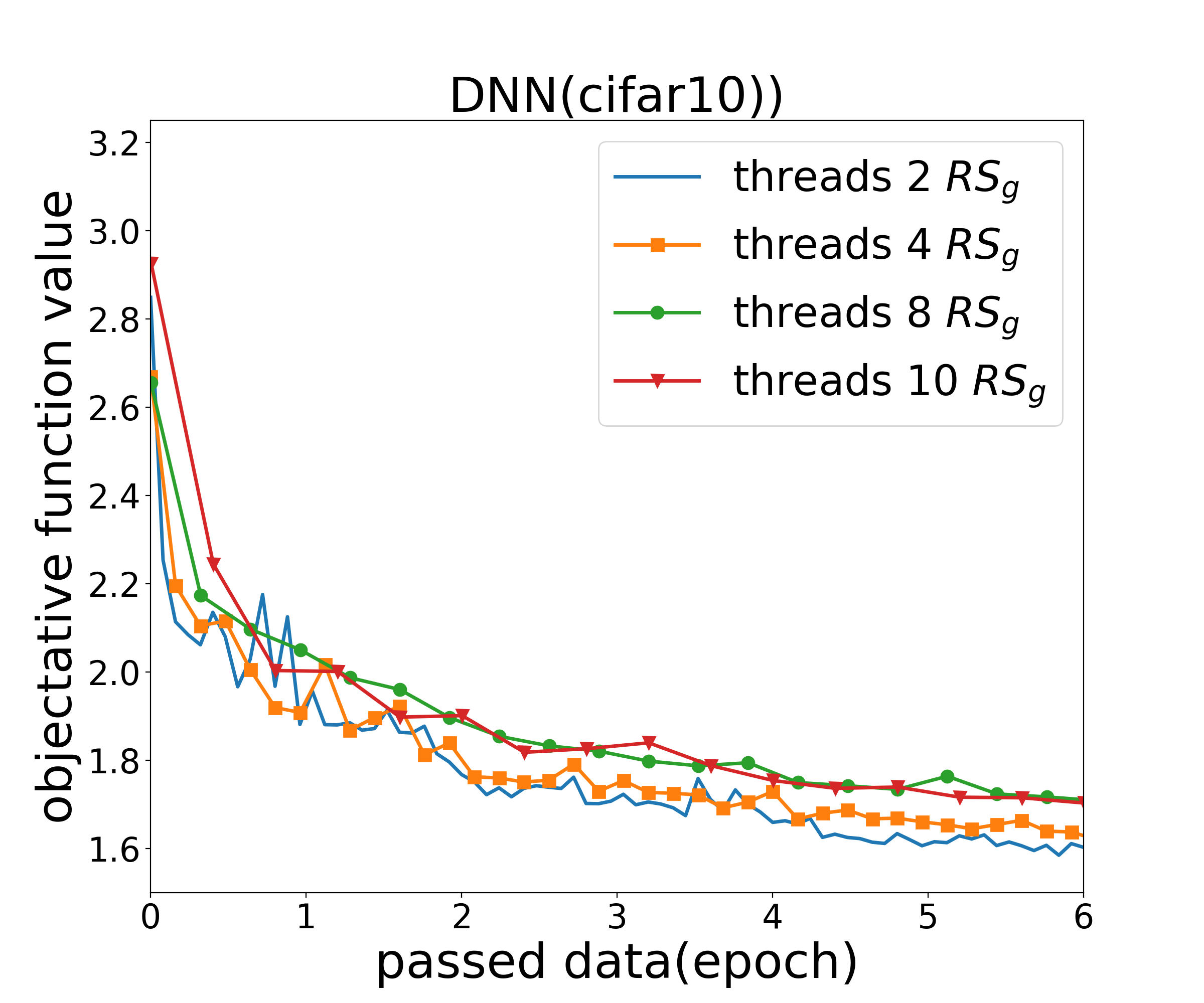}}	
		\subfigure[]{
			\label{fig2c}
			\includegraphics[width=1.3in]{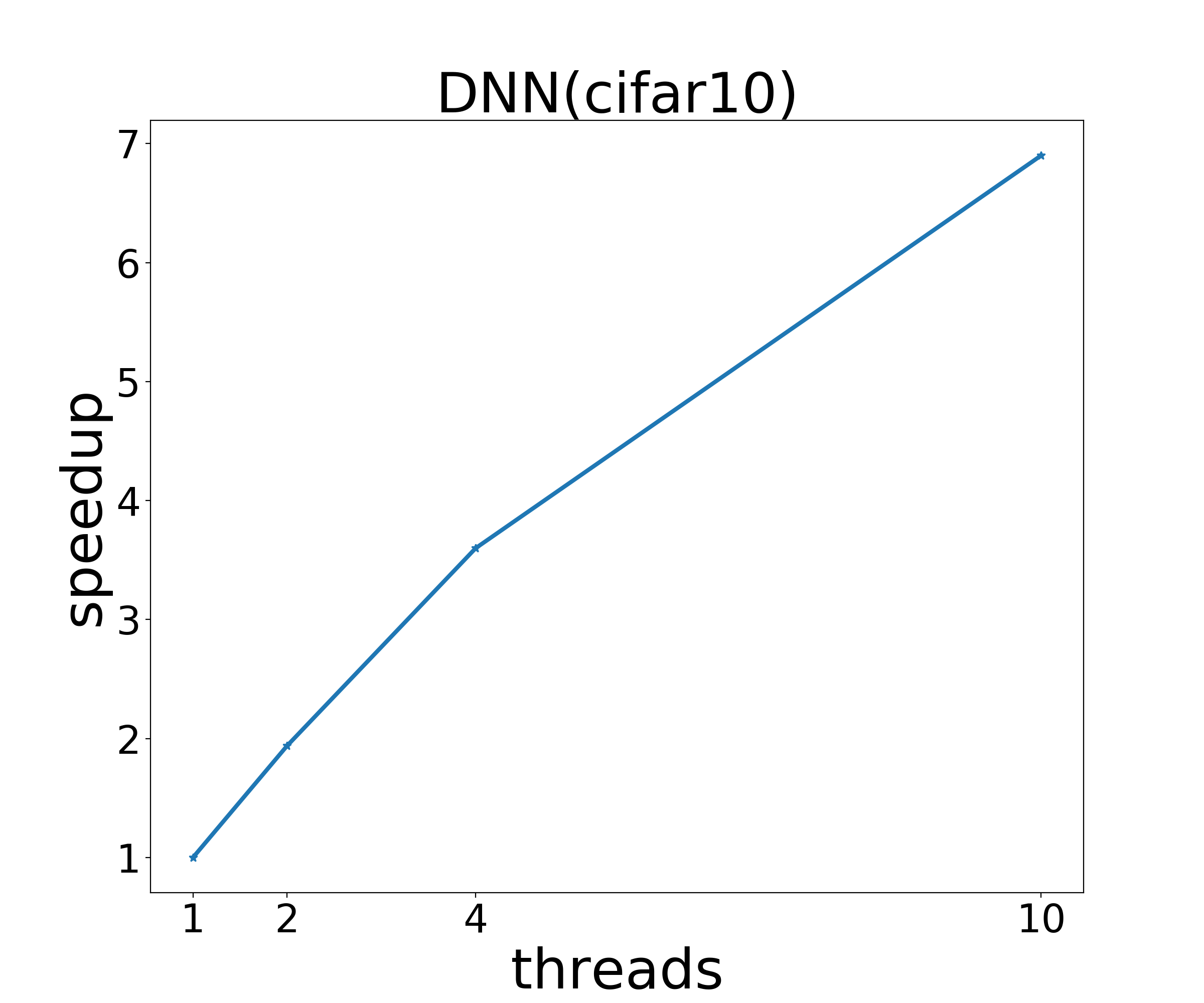}}    	
		\label{Fig2}
		\caption{Speedup for global shuffling in both convex and non-convex cases: Fig(a)(c) are the objective function value curves and (b)(d) are the speedup ratios.}
	\end{figure*}
	\begin{figure*}
		\centering
		\subfigure[]{
			\label{fig3a}
			\includegraphics[width=1.3in]{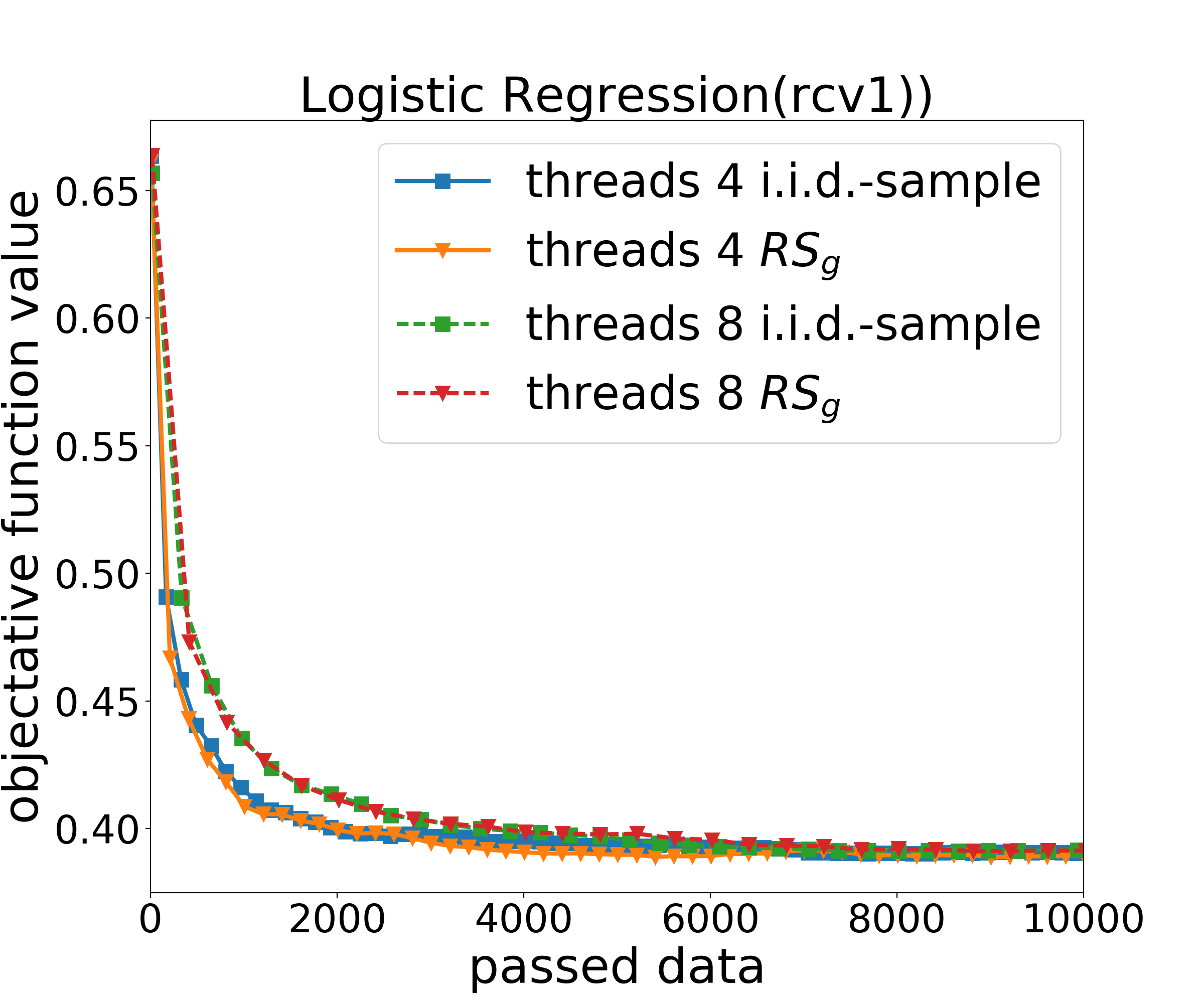}}	   
		\subfigure[]{
			\label{fig3c}
			\includegraphics[width=1.3in]{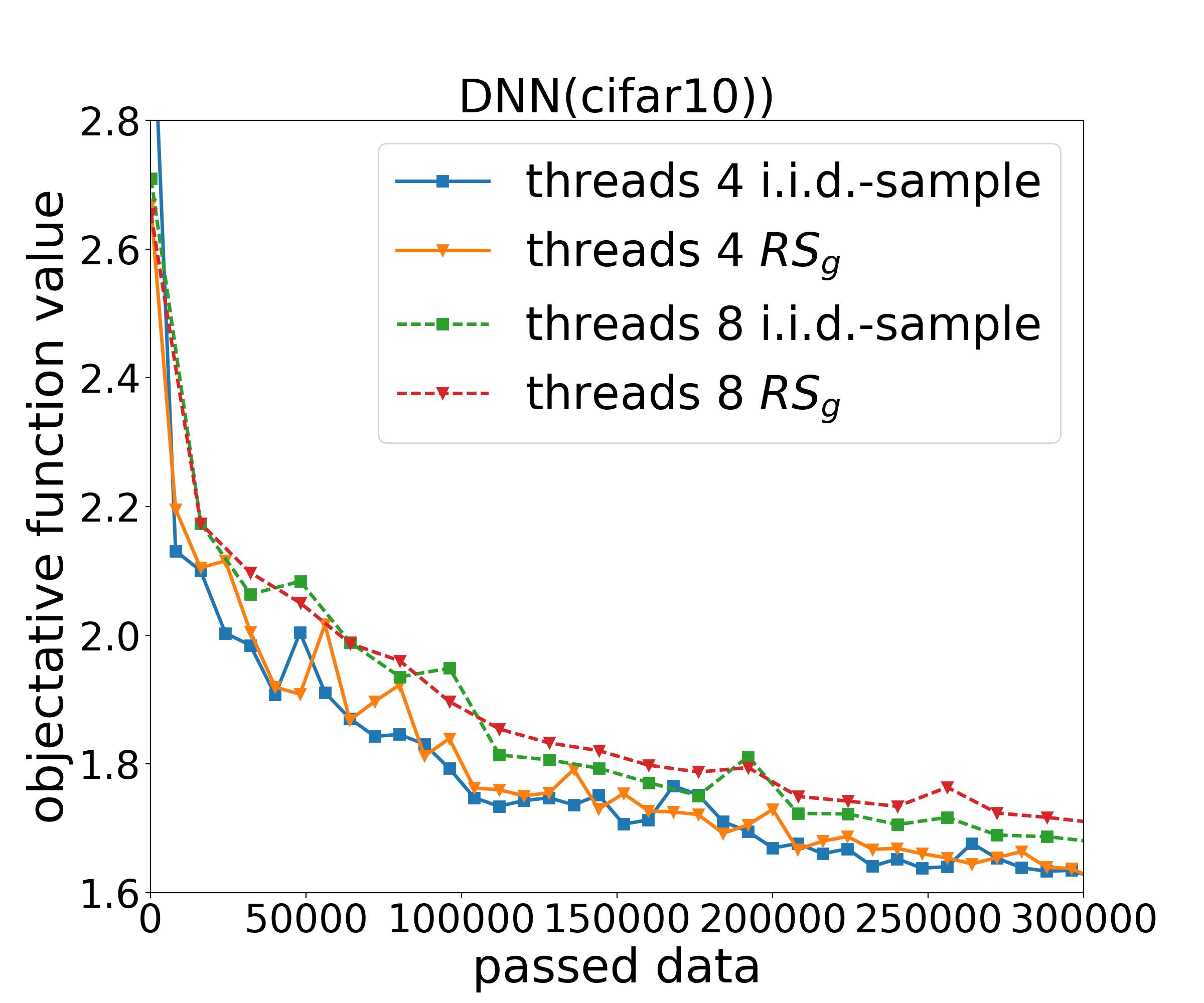}}		    
		\label{Fig3}
		\subfigure[]{
			\label{fig3b}
			\includegraphics[width=1.3in]{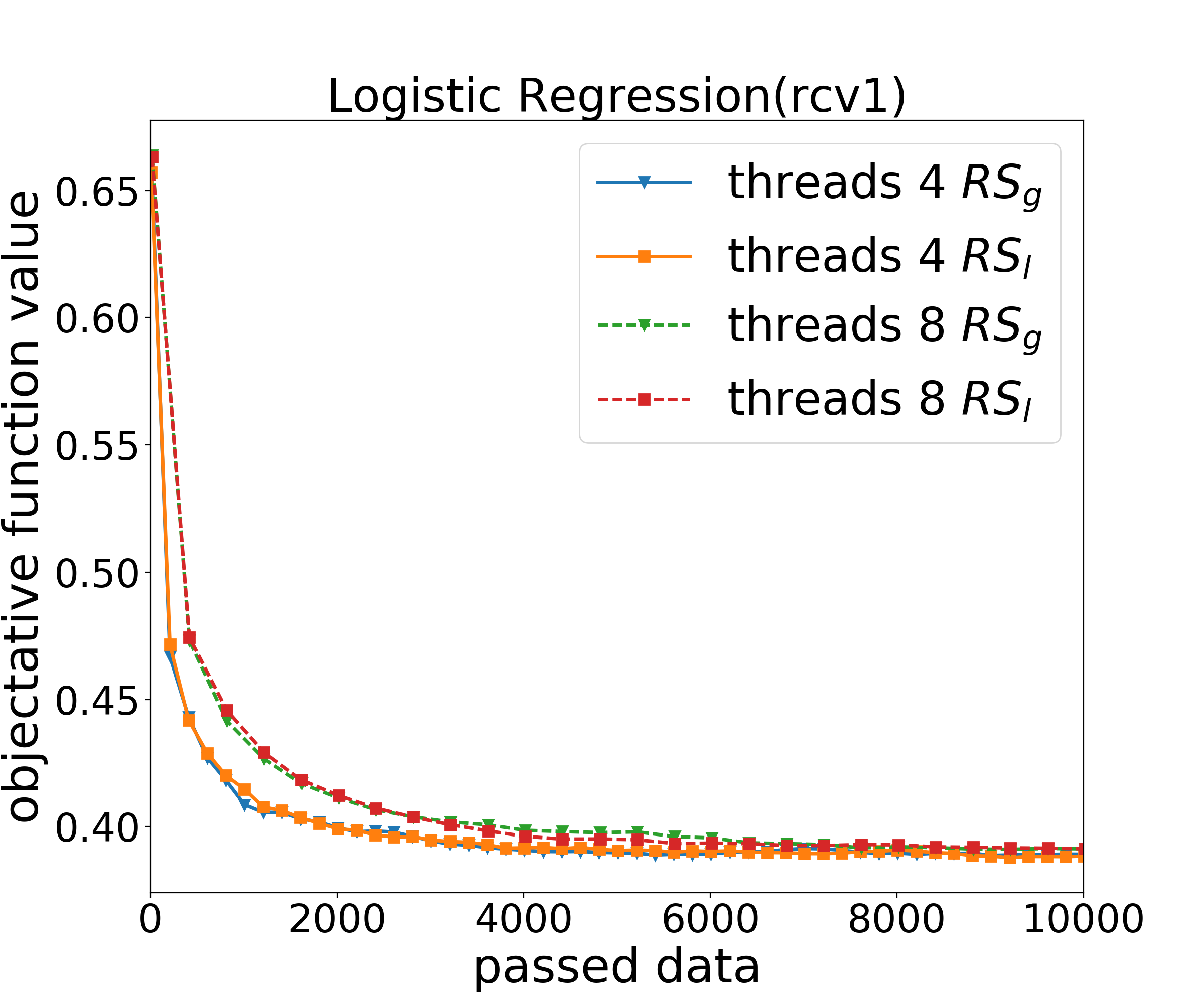}}		   
		\subfigure[]{
			\label{fig3d}
			\includegraphics[width=1.3in]{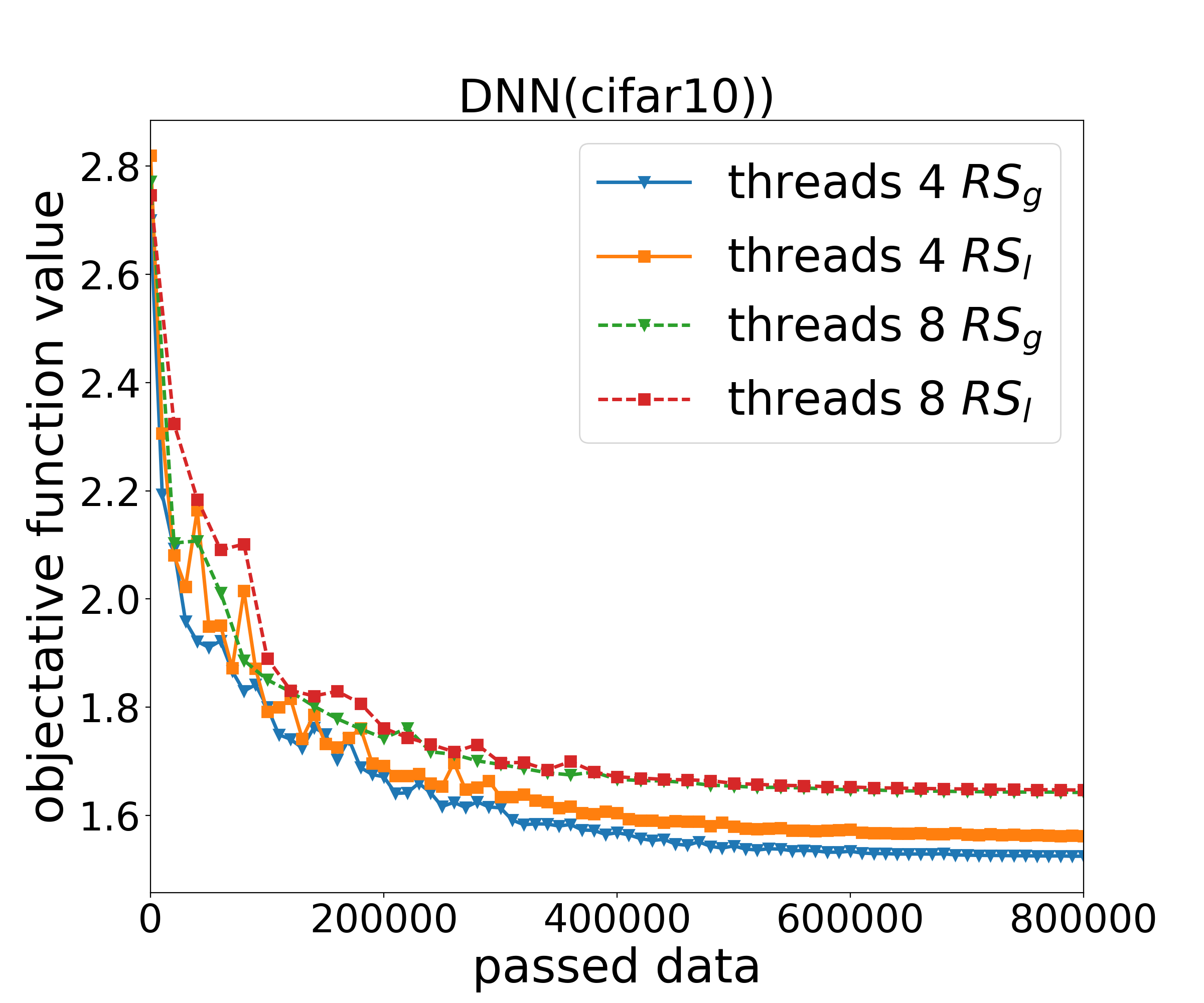}}
		\label{Fig4}
		\caption{Fig (a)(b) are the comparisons of i.i.d sampling and global shuffling; Fig (c)(d) are the comparison of global shuffling and local shuffling.}
	\end{figure*}
	In this section, we report some experimental results to validate our theoretical findings. 
	We conducted experiments on two tasks: logistic regression and fully connected neural networks, whose objective functions are logistic loss and cross-entropy loss respectively. In addition, we used an $L_2$ regularization term with $\lambda=1/\sqrt{n}$ during the training process. The logistic regression experiments are conducted on \emph{RCV1} \cite{chang2011libsvm} dataset. We set the local minibatch size $b=5$ and the learning rate {\small$\eta_t=a_1*\sqrt{1/(t+a_2)}$} with coefficients $a_1$ and $a_2$. We tuned $a_1$ and $a_2$ and report the best results. We used a fully connected neural networks with three hidden layers with $64\times 32\times 16$ hidden nodes in each layer. We trained this neural network on CIFAR-10 \cite{krizhevsky2010convolutional}. The local minibatch size is set to be $b=20$ and the learning rate $\eta=[0.5,0.05,0.005]$. 
	
	We first show the speedups with respect to the number of machines for distributed SGD with global shuffling. For each task, we report two figures. One corresponds to the value of the objective function w.r.t. the total passes of data among all workers. The other corresponds to the speedup ratios w.r.t. the number of workers. For Fig. \ref{fig1c} and \ref{fig2c}, the horizontal axis of each figure corresponds to the number of local workers and the vertical axis corresponds to the speedup when the target training error {\small$(F(w_t)-F(w^*)\leq 10^{-10})$} is achieved. Fig.\ref{fig1a} and \ref{fig1c} show the results for logistic regression. From the figure we can see that the objective function is about to converge when $S=1$. It is clear that the speedup ratio is decreasing as the number of threads becomes larger, which is consistent with our discussion in section \ref{sec3.1}. Fig.1(c) and 1(d) show the results for neural networks. We can see the speedup ratio is nearly linear, which is consistent with the discussion in section \ref{section3.2}.
	
	Then we compare the convergence rate for distributed SGD with i.i.d sampling and global shuffling. Fig.\ref{fig3a} 
	shows the results for logistic regression and Fig.\ref{fig3c} 
	shows the results for neural networks. Both logistic regression and neural networks achieve similar convergence rate, which is consistent with our discussion in section \ref{sec3.1} and section \ref{section3.2}. 
	
	Finally, we compare the convergence rate for global shuffling and local shuffling. Fig.\ref{fig3b} shows the results for logistic regression. We can see that the two shuffling methods achieve similar convergence rate. It is because logistic regression converges with $S=1$. Fig.\ref{fig3d} shows the results for neural networks. From the figure, we can see that the convergence rate for global shuffling is faster than that for local shuffling, which is consistent with the discussions in section \ref{local}. 
	
	\section{Conclusion}
	In this paper, we have conducted a in-depth analysis on the convergence properties of distributed SGD with random shuffling, in both convex and nonconvex cases. Our results show that, in most cases, the convergence rate for random shuffling is comparable to that for i.i.d sampling, and can achieve good speedup ratios. We verify our theoretical findings using the experiments on logistic regression and fully connected neural networks. In the future, we plan to study the convergence rate of other optimization algorithms with random shuffling, and design new data allocation strategy which can have even better the convergence properties. 
	\small
	\bibliography{sampling}

\begin{thebibliography}{10}

\bibitem{GroundHog}
\url{https://github.com/lisa-groundhog/GroundHog/blob/master/groundhog/datasets/TM_dataset.py}.

\bibitem{nyu-dl}
\url{https://github.com/nyu-dl/dl4mt-tutorial/tree/master/data}.

\bibitem{lvsr}
\url{https://github.com/rizar/actor-critic-public/blob/master/lvsr/datasets/__init__.py
  }.

\bibitem{bottou2009curiously}
L{\'e}on Bottou.
\newblock Curiously fast convergence of some stochastic gradient descent
  algorithms.
\newblock In {\em Proceedings of the symposium on learning and data science,
  Paris}, 2009.

\bibitem{bottou2010large}
L{\'e}on Bottou.
\newblock Large-scale machine learning with stochastic gradient descent.
\newblock In {\em Proceedings of COMPSTAT'2010}, pages 177--186. Springer,
  2010.

\bibitem{bousquet2008tradeoffs}
Olivier Bousquet and L{\'e}on Bottou.
\newblock The tradeoffs of large scale learning.
\newblock In {\em Advances in neural information processing systems}, pages
  161--168, 2008.

\bibitem{chang2011libsvm}
Chih-Chung Chang and Chih-Jen Lin.
\newblock Libsvm: a library for support vector machines.
\newblock {\em ACM Transactions on Intelligent Systems and Technology (TIST)},
  2(3):27, 2011.

\bibitem{dean2012large}
Jeffrey Dean, Greg Corrado, Rajat Monga, Kai Chen, Matthieu Devin, Mark Mao,
  Andrew Senior, Paul Tucker, Ke~Yang, Quoc~V Le, et~al.
\newblock Large scale distributed deep networks.
\newblock In {\em Advances in neural information processing systems}, pages
  1223--1231, 2012.

\bibitem{dekel2012optimal}
Ofer Dekel, Ran Gilad-Bachrach, Ohad Shamir, and Lin Xiao.
\newblock Optimal distributed online prediction using mini-batches.
\newblock {\em Journal of Machine Learning Research}, 13(Jan):165--202, 2012.

\bibitem{el2009transductive}
Ran El-Yaniv and Dmitry Pechyony.
\newblock Transductive rademacher complexity and its applications.
\newblock {\em Journal of Artificial Intelligence Research}, 35(1):193, 2009.

\bibitem{fisher1938statistical}
Ronald~Aylmer Fisher, Frank Yates, et~al.
\newblock Statistical tables for biological, agricultural and medical research.
\newblock {\em Statistical tables for biological, agricultural and medical
  research.}, 1938.

\bibitem{ghadimi2013stochastic}
Saeed Ghadimi and Guanghui Lan.
\newblock Stochastic first-and zeroth-order methods for nonconvex stochastic
  programming.
\newblock {\em SIAM Journal on Optimization}, 23(4):2341--2368, 2013.

\bibitem{gurbuzbalaban2015random}
Mert G{\"u}rb{\"u}zbalaban, Asu Ozdaglar, and Pablo Parrilo.
\newblock Why random reshuffling beats stochastic gradient descent.
\newblock {\em arXiv preprint arXiv:1510.08560}, 2015.

\bibitem{johnson2013accelerating}
Rie Johnson and Tong Zhang.
\newblock Accelerating stochastic gradient descent using predictive variance
  reduction.
\newblock In {\em Advances in Neural Information Processing Systems}, pages
  315--323, 2013.

\bibitem{krizhevsky2010convolutional}
Alex Krizhevsky and G~Hinton.
\newblock Convolutional deep belief networks on cifar-10.
\newblock {\em Unpublished manuscript}, 40, 2010.

\bibitem{levin2009markov}
David~Asher Levin, Yuval Peres, and Elizabeth~Lee Wilmer.
\newblock {\em Markov chains and mixing times}.
\newblock American Mathematical Soc., 2009.

\bibitem{li2014efficient}
Mu~Li, Tong Zhang, Yuqiang Chen, and Alexander~J Smola.
\newblock Efficient mini-batch training for stochastic optimization.
\newblock In {\em Proceedings of the 20th ACM SIGKDD international conference
  on Knowledge discovery and data mining}, pages 661--670. ACM, 2014.

\bibitem{nesterov2013introductory}
Yurii Nesterov.
\newblock {\em Introductory lectures on convex optimization: A basic course},
  volume~87.
\newblock Springer Science \& Business Media, 2013.

\bibitem{rakhlin2011making}
Alexander Rakhlin, Ohad Shamir, and Karthik Sridharan.
\newblock Making gradient descent optimal for strongly convex stochastic
  optimization.
\newblock {\em arXiv preprint arXiv:1109.5647}, 2011.

\bibitem{recht2012toward}
Benjamin Recht and Christopher R{\'e}.
\newblock Toward a noncommutative arithmetic-geometric mean inequality:
  Conjectures, case-studies, and consequences.

\bibitem{recht2013parallel}
Benjamin Recht and Christopher R{\'e}.
\newblock Parallel stochastic gradient algorithms for large-scale matrix
  completion.
\newblock {\em Mathematical Programming Computation}, 5(2):201--226, 2013.

\bibitem{rice2006mathematical}
John Rice.
\newblock {\em Mathematical statistics and data analysis}.
\newblock Nelson Education, 2006.

\bibitem{shamir2016without}
Ohad Shamir.
\newblock Without-replacement sampling for stochastic gradient methods:
  Convergence results and application to distributed optimization.
\newblock {\em arXiv preprint arXiv:1603.00570}, 2016.

\bibitem{zinkevich2010parallelized}
Martin Zinkevich, Markus Weimer, Lihong Li, and Alex~J Smola.
\newblock Parallelized stochastic gradient descent.
\newblock In {\em Advances in neural information processing systems}, pages
  2595--2603, 2010.

\end{thebibliography}
	\bibliographystyle{plain}
	
	\section{Appendix}
	\subsection{Relationship between random shuffling and sampling}
	\begin{proposition}\label{prop1}
		If each times of random shuffling is sufficient, and the partition is in order or random, the conditional distribution of the $t$-th minibatch after $RS_g([n];S,M)$ equals that after doing without-replacement sampling, i.e., {\small\begin{align}
			&\mathbb{P}(D_m(t+1)=B_m, m\in[M]|D_m(1),\cdots,D_m(t)\neq B_1,\cdots,B_M, m\in[M])=\frac{1}{T-t},\nonumber
			\end{align}} where $B_m, m\in[M]$ are non-overlapped subsets of the full data and $T=n/Mb$.
	\end{proposition}
	\textit{Proof:} For $n$ instances, the number of different permutations is $n!$. We denote each permutation as $\pi_k([n]),k\in[n!].$ $\sigma([n])$ is a random variable distributed on $\{\pi_1([n]),\cdots,\pi_k([n]),\cdots,\pi_{n!}([n])\}.$ We denote $\pi_k([n])=\{\pi_k(t),t\in[n]\}$ where $\pi_k(t)$ is the $t$-th element in permutation $\pi_k([n])$. If the random shuffling is sufficient, we have $p_k=\mathbb{P}(\sigma([n])=\pi_k)=\frac{1}{n!}, \forall k\in\{1,\cdots,n!\}$.
	We use $I_{[\cdot]}$ to denote the indicator function, then we have
	{\small\begin{align}
		&\mathbb{P}(D_m(t+1)=B_m, m\in[M]|D_m(1),\cdots,D_m(t)\neq B_1,\cdots,B_M, m\in[M])\\
		=&\frac{\mathbb{P}(D_m(t+1)=B_m, m\in[M],D_m(1),\cdots,D_m(t)\neq B_1,\cdots,B_M, m\in[M])}{\mathbb{P}(D_m(1),\cdots,D_m(t)\neq B_1,\cdots,B_M, m\in[M])}\\
		=&\frac{\sum_{k\in[n!]}p_k I_{[\{\pi_k((m-1)n/M+tb+1),\cdots,\pi_k((m-1)n/M+(t+1)b)\}=B_m,m\in [M]]}}{\sum_{k\in[n!]}p_k I_{[\{\pi_k((m-1)n/M+(j-1)b+1),\cdots,\pi_k((m-1)n/M+jb)\}\neq B_1,\cdots B_M,m\in [M],j\in[t]]}}\\
		=&\frac{(n-bM)!}{n!}\cdot\frac{n!}{(n-bM)!(n-tbM)}\\
		=&\frac{1}{n-tbM}=\frac{1}{T-t},
		\end{align}}which is equal to the distribution of without-replacement sampling over $[n]$. $\Box$
	
	\subsection{Relationship between random shuffling and insufficient sampling}
	\begin{definition}
		The shuffling error $\epsilon(A,h,n)$ is defined as the total variation between $u_{\pi([n])}$, which is the uniform distribution on the sets consisted by all the permutations $\pi([n])$, and $v_{\pi([n])(A,h,n)}$, which is the distribution after shuffling the $n$ data using algorithm $A$ with $h$ operators,i.e.,
		\begin{equation}
		\epsilon(A,h,n)=\|u_{\pi([n])}-v_{\pi([n])(A,h,n)}\|_{TV}:=\frac{1}{2}\sum_{\pi_i([n])\in\pi([n])}|u_{\pi_i([n])}-v_{\pi_i([n])}(A,h,n)|.
		\end{equation}
	\end{definition}
	The following proposition establishes the gap between the conditional distribution of $t$-th minibatch under the uniform distribution and the distribution after insufficient shuffling. The difference for the conditional probability can be upper bounded by the shuffling error $\epsilon(A,n)$. 
	
	\begin{proposition}
		Denote {\small$\mathbb{P}_{\cdot|t}(B_m,m\in[M])$} as the conditional probability {\small$\mathbb{P}_\cdot(D_m(t+1)=B_m,m\in[M]|D_m(1),\cdots,D_m(t)\neq B_1,\cdots,B_M, m\in[M])$}. If the shuffling error {\small$\epsilon(A,n)\leq\frac{bM}{n}$}, then for $t+1<T$, we have
		{\small\begin{equation}
			|\mathbb{P}_{v|t}(B_m,m\in[M])-\mathbb{P}_{u|t}(B_m,m\in[M])|\leq\frac{4n\epsilon(A,n)}{n-bMt}.
			\end{equation}}
	\end{proposition} 
	\textit{Proof:}
	Firstly, we have {\small$\mathbb{P}_u(D_m(t+1)=B_m,,m\in[M], D_m(1), \cdots,D_m(t)\neq B_1,\cdots,B_M, m\in[M])=\frac{bM}{n}$} and {\small$\mathbb{P}_u(D_m(1), \cdots,D_m(t)\neq B_1,\cdots,B_M, m\in[M])=\frac{n-tbM}{n}$} because $u$ is the uniform distribution. Let {\small$A_1=\{k:\pi_k((m-1)n/M+tb+1),\cdots,\pi_k((m-1)n/M+(t+1)b)\}=B_m,m\in [M]\}$} and {\small$A_2=\{k:\pi_k((m-1)n/M+(j-1)b+1),\cdots,\pi_k((m-1)n/M+jb)\}\neq B_1,\cdots B_M,m\in [M],j\in[t]\}$} and denote {\small$p_1=\mathbb{P}_v(D_m(t+1)=B_m,,m\in[M], D_m(1), \cdots,D_m(t)\neq B_1,\cdots,B_M, m\in[M])=\sum_{k\in A_1} (p_{v,k})$} and 
	{\small$p_2=\mathbb{P}_v(D_m(1), \cdots,D_m(t)\neq B_1,\cdots,B_M, m\in[M])
		=\sum_{k\in A_2}(p_{v,k})$}. By the definition of $\|u-v\|_{TV}$, we have {\small$\left|p_1-\frac{1}{n}\right|\leq 2\epsilon(A,n)\leq\frac{bM}{n}$} and {\small$\left|p_2-\frac{n-tbM}{n}\right|\leq2\epsilon(A,n)\leq\frac{bM}{n}$}. Then we have for $t+1<n$,
	\begin{align}
	|\mathbb{P}_{v|t}(i)-\mathbb{P}_{u|t}(i)|\leq&\Big|\frac{p_1}{p_2}-\frac{bM/n}{(n-tbM)/n}\Big|\\
	=&\Big|\frac{p_1\cdot\frac{n-tbM}{n}-p_2\cdot\frac{bM}{n} }{\frac{n-tbM}{n}\cdot p_2}\Big|\\
	\leq&\Big|\frac{p_1-\frac{bM}{n}}{p_2}\Big|+\frac{bM}{n-tbM}\Big|\frac{p_2-\frac{n-tbM}{n}}{p_2}\Big|\\
	\leq&\frac{n-tbM+bM}{n-tbM}\cdot\frac{2\epsilon(A,n)}{p_2}\\
	\leq&\frac{n-tbM+bM}{n-tbM}\cdot\frac{2\epsilon(A,n)}{\frac{n-tbM}{n}-\frac{bM}{n}}\label{ineq7}\\
	\leq&\frac{4n\epsilon(A,n)}{n-tbM},
	\end{align}where inequality \ref{ineq7} is based on $\left|p_2-\frac{n-tbM}{n}\right|\leq\frac{bM}{n}$. 
	
	\subsection{Transductive Rademacher Complexity }
	\begin{definition}
		Let $\mathcal{V}$ be a set of vectors $\textbf{v}=(v_1,\cdots,v_n)$ in $\mathbb{R}^n$. Let $s,u$ be positive integers such that $s+u=n$, and denote $p:=\frac{su}{(s+u)^2}\in(0,0.5)$. We define the transductive Rademacher Complexity $\mathcal{R}_{s,u}(\mathcal{V})$ as
		{\small\begin{equation} \mathcal{R}_{s,u}(\mathcal{V})=\left(\frac{1}{s}+\frac{1}{u}\right)\cdot\mathbb{E}_{r_1,\cdots,r_n}\left(\sup_{\textbf{v}\in\mathcal{V}}\sum_{i=1}^nr_iv_i\right),
			\end{equation}}
		where $r_1,\cdots,r_n$ are i.i.d. random variables such that $r_i=1$ with probability $p$, $r_i=-1$ with probability $p$ and $r_i=0$ with probability $1-p$.
	\end{definition}
	The following lemma in \cite{shamir2016without} provides an upper bound for the transductive Rademacher complexity.
	\begin{lemma}\label{lemma2.2}
		Let $\mathcal{V}=\{v_i,i\in[n];v_i\leq B\}$, we have
		{\small$ \mathcal{R}_{s,u}(\mathcal{V})\leq\sqrt{2}\left(\frac{1}{\sqrt{s}}+\frac{1}{\sqrt{u}}\right)B.
			$}
	\end{lemma}
	In order to prove the convergence rates of distributed SGD with random shuffling, we first show the following lemmas which extend Lemma 1 and Corollary 2 in \cite{shamir2016without} into the mini-batch case.
	\begin{lemma}\label{lemma3.1}
		Let $\sigma$ be a random permutation over $\{1,\cdots,n\}$ chosen uniformly at random variables conditioned on $\sigma(1),\cdots,\sigma(tb)$\footnote{For simplicity, we denote $\sigma([n])$ as $\sigma$ and $\sigma([n])_t$ as $\sigma(t)$.}, which are independent of $\sigma(tb+1),\cdots,\sigma(n)$. Let $s_{a:b}=\frac{1}{b+1-a}\sum_{i=a}^bs_i$. Then, we have $\forall t>1$,
		\begin{equation}
		\mathbb{E}\left[\frac{1}{n}\sum_{i=1}^ns_i-\frac{1}{b}\sum_{j=1}^{b}s_{\sigma(tb+j)}\right]=\frac{tb}{n}\cdot\mathbb{E}[s_{1:tb}-s_{tb+1:n}]
		\end{equation}
	\end{lemma} 
	\textbf{Proof:}
	{\small\begin{align}
		\mathbb{E}\left[\frac{1}{n}\sum_{i=1}^ns_i-\frac{1}{b}\sum_{j=1}^{b}s_{\sigma(t+j)}\right]
		=&\mathbb{E}\left[\frac{1}{n}\sum_{i=1}^ns_i-\frac{1}{C_{n-tb}^b}\cdot\frac{1}{b}\cdot C_{n-tb+1}^{b-1}\sum_{i=tb+1}^{n}s_{i}\right]\\
		=&\frac{tb}{n}\cdot\mathbb{E}[s_{1:tb}-s_{tb+1:n}]
		\end{align}}
	\begin{lemma}\label{lemma3.2}
		Suppose $S\subset[-B,B]^n$ for some $B>0$. Let $\sigma$ be a random permutation over $\{1,\cdots,n\}$. Then we have
		{\small\begin{align*}
			\mathbb{E}(\sup_{s\in\mathcal{S}}{(s_{1:tb}-s_{tb+1:n})})&\leq\mathcal{R}_{tb,n-tb}(\mathcal{S})+12B(\frac{1}{\sqrt{tb}}+\frac{1}{\sqrt{n-tb}})\\
			\sqrt{\mathbb{E}\left[\sup_{s\in\mathcal{S}}{(s_{1:tb}-s_{tb+1:n})}\right]^2}&\leq\sqrt{2}\mathcal{R}_{tb,n-tb}(\mathcal{S})+12\sqrt{2}B(\frac{1}{\sqrt{tb}}+\frac{1}{\sqrt{n-tb}}).
			\end{align*}}
	\end{lemma}
	\subsection{Convergence rate of global shuffling}
	\subsubsection{Strongly-convex case}
	In order to prove the main theorem, we firstly prove a lemma to show the convergence rate for SGD with global shuffling $RS_{g}([n],1,1)$. 
	
	\begin{lemma} Suppose that Assumption 1 holds. Distributed SGD with global random  shuffling $RS_{g}([n],1,1)$ and learning rate $\eta_t=\frac{4}{\mu t}$ for strongly convex and smooth objective function has convergence rate
		{\small\begin{align}\label{eq17}
			&\mathbb{E}\|w_{t}-w^*\|^2\leq \mathcal{O}\left(\min\left\{\frac{\kappa^2\log{t}}{t^2},\frac{1}{t}\right\}+\frac{\log{t}}{bt}\right).
			\end{align}}
	\end{lemma}
	
	\textit{Proof:}
	
	Let $A_t=w_t-w^*$ and {\small$g_{D(t)}(w)=\frac{1}{b}\sum_{i\in b_t}\nabla f_{i}(w)$}, where $b_t$ is the $t$-th minibatch.
	
	Firstly we decompose the term by using Assumption 1 as
	{\small\begin{align}
		&\mathbb{E}\|A_{t+1}\|^2 \label{eq10}\\
		=&\|A_{t}\|^2-2\eta_t\mathbb{E}\langle g_{D(t)}(w_t), A_{t}\rangle+\eta_t^2\mathbb{E}\|g_{D(t)}(w_t)\|^2\\
		=&\|A_{t}\|^2-2\eta_t\langle\nabla F(w_t), A_{t}\rangle+2\eta_t\mathbb{E}\langle \nabla F(w_t)-g_{D(t)}, A_{t}\rangle+3\eta_t^2\mathbb{E}\|g_{D(t)}-\nabla f_{tb+1:n}(w_t)\|^2 \label{eq29}\\
		&+3\eta_t^2\mathbb{E}\|\nabla f_{tb+1:n}(w_t)-\nabla F(w_t)\|^2+3\eta_t^2\|\nabla F(w_t)\|^2 \label{eq30}\\
		\leq&\|A_t\|^2-2\eta_t\langle\nabla F(w_t),A_{t}\rangle+3\eta_t^2\|\nabla F(w_t)\|^2 +3\eta_t^2\mathbb{E}\|g_{D(t)}-\nabla f_{tb+1:n}(w_t)\|^2 \label{eq16}\\
		&+2\eta_t\mathbb{E}\langle \nabla F(w_t)-g_{D(t)},A_{t}\rangle+3\eta_t^2\mathbb{E}\|\nabla f_{tb+1:n}(w_t)-\nabla F(w_t)\|^2 \label{eq18}
		\end{align}}Using strongly convex assumption, we have 
	{\small\begin{align}
		-\langle\nabla F(w_t),A_{t}\rangle&\leq-(F(w_t)-F(w^*)+\frac{\mu}{2}\|w_t-w^*\|^2)\leq-\mu\|w_t-w^*\|^2. \label{eq20}
		\end{align}}
	For the first term of Eq.(\ref{eq18}), using the AM-GM inequality, we have
	{\small\begin{align}
		\mathbb{E}\langle\nabla F(w_t)-g_{D(t)},A_{t}\rangle\leq\frac{\mu}{2}\|A_{t}\|^2+\frac{1}{2\mu}\mathbb{E}\|\nabla F(w_t)-\nabla f_{tb+1:n}(w_t)\|^2 \label{eq21}
		\end{align}}Using Lemma \ref{lemma3.1} and Lemma \ref{lemma3.2}, we can bound the second term in above equation and Eq.(\ref{eq18}) as follows,
	{\small\begin{align}
		\mathbb{E}\|\nabla F(w_t)-g_{D(t)}\|^2&=\mathbb{E}\|\nabla f_{1:bt}-\nabla f_{bt+1:n}\|^2\\
		&\leq 2(2+12\sqrt{2})^2B^2\cdot\frac{b^2t^2}{n^2}\left(\frac{1}{tb}+\frac{1}{n-tb}\right) \label{eq23}
		\end{align}}For the last term in Eq.(\ref{eq18}), it is bounded by using Theorem B in page 208 \cite{rice2006mathematical} as
	{\small\begin{align}
		\mathbb{E}\|g_{D(t)}-\nabla F(w_t)\|^2\leq\frac{(n-(t+1)b)B^2}{b(n-tb)}\leq\frac{B^2}{b}.\label{eq24}
		\end{align}}
	
	Putting Eq.(\ref{eq20}), Eq.(\ref{eq21}), Eq.(\ref{eq24}) in Eq.(\ref{eq10}), we get
	{\small\begin{align}
		\mathbb{E}\|A_{t}\|^2
		\leq(1-\frac{\mu\eta_t}{2})\|A_{t}\|^2+3\eta_t^2G^2+\frac{3\eta_t^2B^2}{b}+(3\eta_t^2+\frac{\eta_t}{\mu})\frac{b^2t^2}{n^2}\left(\frac{1}{tb}+\frac{1}{n-tb}\right)584B^2
		\end{align}}
	
	If we set $\eta_t=\frac{4}{\mu t}$, we can get
	{\small\begin{align}
		\mathbb{E}\|A_{t}\|^2
		\leq&(1-\frac{2}{t})\|A_{t}\|^2+\frac{48G^2}{\mu^2t^2}+\frac{48B^2}{b\mu^2t^2}+584B^2\left(\frac{48}{\mu^2t^2}+\frac{4}{\mu^2t}\right)\frac{b^2t^2}{n^2}\left(\frac{1}{tb}+\frac{1}{n-tb}\right)\\
		\leq&(1-\frac{2}{t})\|A_{t}\|^2+\frac{48G^2}{\mu^2t^2}+\frac{48B^2}{b\mu^2t^2}+584B^2\left(\frac{48}{\mu^2t}+\frac{4}{\mu^2}\right)\left(\frac{1}{n(T-t)}\right)\\
		\end{align}}We have
	{\small\begin{align}
		\frac{1}{n}\left(\frac{1}{t(T-t)}+\sum_{j=3}^t\frac{j-2}{j}\frac{1}{(j-1)(T-(j-1))}\right)
		\leq\frac{1}{nt(t-1)}\int_1^t\frac{x}{T-x}\mathrm{d}x\leq \frac{T\log{t}}{nt^2}=\frac{\log{t}}{bt^2}
		\end{align}}and
	{\small\begin{align}
		\frac{1}{n}\left(	\frac{1}{T-t}+\sum_{j=3}^t\frac{j-2}{j}\frac{1}{T-(j-1)}\right)
		\leq\frac{1}{nt}\int_1^t\frac{x}{T-x}\mathrm{d}x\leq\frac{\log{t}}{bt}.
		\end{align}}By induction we can get 
	{\small\begin{align}
		&\mathbb{E}\|w_{t}-w^*\|^2\leq \mathcal{O}\left(\frac{G^2}{t}+\frac{B^2}{tb}+\frac{\log{t}}{tb}\right).
		\end{align}}By the smoothness assumption, we have {\small$\|\nabla F(w_t)\|^2\leq \rho^2\|w_t-w^*\|^2\leq\mathcal{O}\left(\frac{\kappa^2}{t}\right)$}, which is a decreasing sequence. 
	We can further improve the bound by exchange $\|\nabla F(w_t)\|^2\leq G^2$ into $\|\nabla F(w_t)\|^2\leq\mathcal{O}\left(\frac{\kappa^2}{t}\right)$. Then we can get 
	{\small\begin{align}
		\mathbb{E}\|A_{t}\|^2\leq&(1-\frac{2}{t})\|A_{t}\|^2+\mathcal{O}\left(\frac{G^2\kappa^2}{\mu^2t^3}\right)+\frac{48B^2}{b\mu^2t^2}+584B^2\left(\frac{48}{\mu^2t}+\frac{4}{\mu^2}\right)\left(\frac{1}{n(T-t)}\right).
		\end{align}}By induction, we have
	{\small\begin{align}
		&\mathbb{E}\|w_{t}-w^*\|^2\leq \mathcal{O}\left(\min\left\{\frac{\kappa^2\log{t}}{t^2},\frac{1}{t}\right\}+\frac{\log{t}}{bt}\right).
		\end{align}}
	$\Box$
	
	It is trivial to extend the results in $RS_{g}([n],1,M)$ case by replace $b$ with $Mb$. Based on the above lemma, we can get the following results in $RS_g([n],S,M)$ case by applying induction from $s=1$ to $S$.
	
	\begin{theorem}
		Suppose the objective function is strongly convex and smooth, and Assumption 1 holds. Then distributed SGD with global shuffling {\small$RS_g([n],S,M)$} 
		and learning rate {\small$\eta_t^s= \frac{2}{\mu((s-1)T+t)}$} where {\small$T=\frac{n}{bM}$}, has the following convergence rate :
		{\small\begin{align*}
			&\mathbb{E}\|w^{S}-w^*\|^2\leq\mathcal{O}\left(\min\left\{\frac{bM}{Sn},\frac{\kappa^2(bM)^2\log{Sn}}{(Sn)^2}+\frac{\kappa^2bM\log{n}}{Sn^2}\right\}+\frac{\log{n}}{n}\right).
			\end{align*}}
	\end{theorem}
	\textit{Proof:}
	At the $s+1$ stage, the learning rate $\eta_t^s=\frac{4}{\mu((s-1)T+t)}$,
	we have
	{\small\begin{align*}
		&\mathbb{E}\|w_{T}^{S}-w^*\|^2\\
		\leq&\left(1-\frac{2}{ST}\right)\|w_{T-1}^S-w^*\|^2+\left(\frac{48G^2}{\mu^2(ST)^2}\right)+\frac{48B^2}{bM\mu^2(ST)^2} \\
		&+584B^2\left(\frac{48}{\mu^2(ST)^2}+\frac{4}{\mu^2(ST)}\right)\left(\frac{T-1}{n(T-(T-1))}\right)\\
		\leq&\frac{((S-1)T-1)((S-1)T)}{(ST-1)(ST)}\mathbb{E}\|w_T^{S-1}-w^*\|^2+\frac{48(B^2/bM+G^2)T}{\mu^2(ST)(ST-1)}\\
		&+584B^2\left(\frac{48\log{T}}{\mu^2(ST-1)(ST)b}+\frac{4(S-1)T\log{T}}{\mu^2(ST-1)(ST)bM}\right)
		\end{align*}}
	Thus by induction, we have 
	{\small\begin{align*}
		&\mathbb{E}\|w_T^S-w^*\|^2\\
		\leq&\frac{(S-1)^2}{S^2}\mathbb{E}\|w_T^{S-1}-w^*\|^2+A\frac{1}{S(ST-1)}+B\frac{\log{T}}{(ST-1)(ST)bM}+C\frac{(S-1)\log{T}}{(ST-1)SbM}\\
		\leq&\mathcal{O}\left(\min\left\{\frac{\kappa^2\log{T}}{S^2T^2},\frac{1}{S^2T}\right\}+\frac{\log{T}}{bMTS^2}\right)+A\frac{1}{ST}+B\frac{\log{T}}{bMST^2}+C\frac{\log{T}}{bMT},
		\end{align*}}where $A=\frac{48(B^2/bM+G^2)}{\mu^2}$, $B=\frac{584\cdot48 B^2}{\mu^2}$ and $C=\frac{584\cdot4 B^2}{\mu^2}$.
	Thus we have 
	{\small\begin{align}\label{eq54}
		\mathbb{E}\|\nabla F(w_T^S)\|^2\leq \mathcal{O}\left(\frac{\kappa^2}{ST}+\frac{\kappa^2\log{T}}{n}\right)
		\end{align}}
	Then we use Ineq.(\ref{eq54}) to refine the term $\frac{1}{ST}$, we have 
	{\small\begin{align}
		&\mathbb{E}\|w_T^S-w^*\|^2\\
		\leq&\mathcal{O}\left(\min\left\{\frac{\kappa^2\log{T}}{S^2T^2},\frac{1}{S^2T}\right\}+\frac{1}{nS}+\frac{\log{T}}{nS^2}+\frac{\log{T}}{STn}+\frac{\log{T}}{n}\right)+\mathcal{O}\left(\min\left\{\frac{1}{ST},\frac{\kappa^2}{ST^2}+\frac{\kappa^2}{STn}\right\}\right)\\
		\leq&\mathcal{O}\left(\min\left\{\frac{1}{ST},\frac{\kappa^2\log{ST}}{S^2T^2}+\frac{\kappa^2\log{T}}{STn}\right\}+\frac{\log{T}}{n}\right)
		\end{align}}Since we denote $w_T^S$ as $w^S$ and $n=bMT$, we have the results in the theorem. $\Box$
	
	Based on the above theorem, we have the following corollary.
	\begin{corollary}
		If {\small$S\leq \frac{bM\kappa^2}{n}$}, the convergence rate of distributed SGD with global shuffling {\small$RS_g([n],S,M)$} is comparable with distributed SGD with i.i.d sampling.
		If {\small$S\geq bM\max\{1,\frac{\kappa^2}{n}\}$}, distributed SGD with global shuffling {\small$RS_g([n],S,M)$} achieves at least linear speedup compared with the sequential SGD with {\small$RS_g([n],S,1)$}.
	\end{corollary}
	\textit{Proof:} 
	
	If {\small$S\leq \frac{bM\kappa^2}{n}$}, the term {\small$\mathcal{O}(\frac{1}{ST})$} will dominate the bound. Therefore, the convergence rate of distributed SGD with global shuffling is comparable with that with with-replacement sampling (See Theorem \ref{with}.). 
	
	For distributed SGD, we assume that each local worker has the same computation efficiency and the communication cost is negligible. If we want to achieve linear speedup, the term {\small$\frac{1}{n}$} should dominate the bound. This can be guaranteed, if {\small$S\geq bM\max\{1,\frac{\kappa^2}{n}\}$}. 
	\subsubsection{Convex case}
	\begin{theorem}
		Suppose the objective function is convex and $L$-Lipschitz, and Assumption 1 holds. Then distributed SGD with global shuffling {\small$RS_g([n],S,M)$} and learning rate {\small$\eta_t^s=\sqrt{\frac{L}{((s-1)T+t)}}$} where $T=\frac{n}{bM}$, has the following convergence rate:
		{\small$
			\mathbb{E}F(\bar{w}^S)-F(w^*)\leq\mathcal{O}\left\{\frac{1}{\sqrt{nS}}+\frac{Mb}{nS}+\sqrt{\frac{1}{n}}\right\}.
			$}
	\end{theorem}
	\textit{Proof:} Following the proof of Theorem 1 in \cite{shamir2016without}, we have
	{\small\begin{align}
		&\mathbb{E}\left[\frac{1}{TS}\sum_{s=1}^S\sum_{t=1}^TF(w_t^s)-F(w^*)\right]\\
		=&\mathbb{E}\left[\frac{1}{TS}\sum_{s=1}^S\sum_{t=1}^T\left(F(w_t^s)-\frac{1}{M}\sum_{m=1}^Mf_{D_m^s(t)}(w_t^s)\right)\right]+\mathbb{E}\left[\frac{1}{TS}\sum_{s=1}^S\sum_{t=1}^T(\frac{1}{M}\sum_{m=1}^M(f_{D_m^s(t)}(w_t^s)-f_{D_m^s(t)}(w^*))\right]\label{eq420}.
		\end{align}}
	The upper bound for the second term in Ineq.(\ref{eq420}) comes from the regrets for mini-batch sgd \cite{dekel2012optimal}, i.e.,
	{\small\begin{align}
		\mathbb{E}\left[\frac{1}{TS}\sum_{s=1}^S\sum_{t=1}^T(\frac{1}{M}\sum_{m=1}^M(f_{D_m^s(t)}(w_t^s)-f_{D_m^s(t)}(w^*))\right]\leq\mathcal{O}\left(\sqrt{\frac{1}{bMTS}}+\frac{1}{TS}\right)=\mathcal{O}\left(\sqrt{\frac{1}{nS}}+\frac{bM}{nS}\right).\label{eq421}
		\end{align}}
	By using Lemma \ref{lemma2.2}, Lemma \ref{lemma3.1} and Lemma \ref{lemma3.2}, we have the upper bound for the first term in Ineq.(\ref{eq420}) is
	{\small\begin{align}
		&\mathbb{E}\left[\frac{1}{TS}\sum_{s=1}^S\sum_{t=1}^T\left(F(w_t^s)-\frac{1}{M}\sum_{m=1}^Mf_{D_m^s(t)}(w_t^s)\right)\right]\\
		\leq &\frac{(2+12\sqrt{2})B}{TS}\sum_{s=1}^S\int_{t=1}^T\frac{tMb}{n}\left(\frac{1}{\sqrt{tbM}}+\frac{1}{\sqrt{n-tbM}}\right)dt\\
		\leq&\mathcal{O}\left(\frac{1}{\sqrt{n}}\right)\label{eq424}
		\end{align}}
	Combining Ineq.(\ref{eq421}) and Ineq.(\ref{eq424}), we can get the results.
	
	\begin{corollary}
		If {\small$S\leq \frac{Mb}{\sqrt{n}}$}, the convergence rate of distributed SGD with global shuffling {\small$RS_g([n],S,M)$} is comparable with distributed SGD with i.i.d sampling.
		If {\small$S>\frac{Mb}{\sqrt{n}}$}, distributed SGD with global shuffling {\small$RS_g([n],S,M)$} achieves linear speedup compared with the sequential SGD with {\small$RS_g([n],S,1)$}.
	\end{corollary}
	\textit{Proof:}
	
	Based on the above theorem, we can see that if the number of epochs {\small$S\leq\frac{Mb}{\sqrt{n}}$}, the extra term {\small$\sqrt{\frac{1}{n}}$} will not dominate the bound. Thus distributed SGD with global shuffling is comparable with with-replacement sampling. 
	
	If $S>\frac{Mb}{\sqrt{n}}$, the term $\frac{Mb}{nS}$ will not dominate the bound and then distributed SGD with global shuffling achieves linear speedup.

	\subsubsection{Nonconvex case}
	\begin{theorem}
		Suppose the objective function is non-convex and $\rho$-smooth, and Assumption 1 holds. By setting {\small$\eta=\min\left\{\frac{1}{\sqrt{ST}}\cdot\sqrt{\frac{2(F(w_0^1)-F(w^*))}{\frac{3\rho B^2}{bM}\left(1+\frac{584\log{T}}{T}\right)}},\frac{1}{6\rho}\right\}$} where $T=\frac{n}{bM}$, distributed SGD with global shuffling {\small$RS_g([n],S,M)$} has the following convergence rate:
		{\small$
			\frac{1}{TS}\left(\sum_{s=1}^S\sum_{t=1}^T\|\nabla F(w_t^s)\|^2\right)\leq\mathcal{O}\left(\sqrt{\frac{(F(w_0^1)-F(w^*))\rho}{Sn}}+\frac{\log{(n)}}{n}\right).
			$}
	\end{theorem}	
	\textit{Proof:} We simply proof the case $M=1$ and the results can be extend to any $M>1$ by replacing $b$ to $Mb$. Conditioned on the previous information $w^1,\cdots, w^{S-1},D_m^S, m\in[M]$, using the smooth condition, we have
	{\small\begin{align}
		&\mathbb{E}F(w_{t}^S) \label{eq50}\\
		\leq&F(w_{t-1}^S)+\mathbb{E}\langle\nabla F(w_{t-1}^S),w_{t}^S-w_{t-1}^S\rangle+\frac{\rho}{2}\mathbb{E}\|w_{t}^S-w_{t-1}^S\|^2 \nonumber\\
		=&F(w_{t-1}^S)-\eta_t\mathbb{E}\langle\nabla F(w_{t-1}^S),g_{D^S(t)}(w_{t-1}^S)\rangle+\frac{\rho\eta_t^2}{2}\mathbb{E}\|g_{D^S(t)}(w_{t-1}^S)\|^2 \nonumber\\
		\leq&F(w_{t-1}^S)-\eta_t\|\nabla F(w_{t-1}^S)\|^2-\eta_t\mathbb{E}\langle\nabla F(w_{t-1}^S),g_{D^S(t)}(w_{t-1}^S)-\nabla F(w_{t-1}^S)\rangle\\
		&+\frac{3\rho\eta_t^2}{2}\mathbb{E}\|g_{D^S(t)}(w_{t-1}^S)-\nabla f_{tb:n-tb}(w_{t-1}^S)\|^2\\
		&+\frac{3\rho\eta_t^2}{2}\mathbb{E}\|\nabla f_{tb:n-tb}(w_{t-1}^S)-\nabla F(w_{t-1}^S)\|^2+\frac{3\rho\eta_t^2}{2}\|\nabla F(w_{t-1}^S)\|^2 \nonumber\\
		\leq&F(w_{t-1}^S)-\eta_t\|\nabla F(w_{t-1}^S)\|^2+\frac{\eta_t}{2}\|\nabla F(w_{t-1}^S)\|^2+\frac{\eta_t}{2}\mathbb{E}\|g_{D^S(t)}(w_{t-1}^S)-\nabla F(w_{t-1}^S)\|^2 \nonumber\\
		&+\frac{3\rho\eta_t^2}{2}\mathbb{E}\|g_{D^S(t)}(w_{t-1}^S)-\nabla f_{tb:n-tb}(w_{t-1}^S)\|^2+\frac{3\rho\eta_t^2}{2}\mathbb{E}\|\nabla f_{tb:n-tb}(w_{t-1}^S)-\nabla F(w_{t-1}^S)\|^2\\
		&+\frac{3\rho\eta_t^2}{2}\|\nabla F(w_{t-1}^S)\|^2\nonumber
		\end{align}}Using Eq.(\ref{eq23}), we have {\small
		$\mathbb{E}\|g_{D^S(t)}(w_{t-1}^S)-\nabla F(w_{t-1}^S)\|^2$} and
	{\small$\mathbb{E}\|\nabla f_{tb:n-tb}(w_{t-1}^S)-\nabla F(w_t)\|^2$
	}are all upper bounded by {\small$2(2+12\sqrt{2})^2B^2\cdot\frac{b^2t^2}{n^2}\left(\frac{1}{tb}+\frac{1}{n-tb}\right)$}, which is equal to $\frac{2(2+12\sqrt{2})^2B^2 t}{n(T-t)}$.
	
	Rearrange Eq.(\ref{eq50}), we can get
	{\small\begin{align}
		&\left(\frac{\eta_t}{2}-\frac{3\rho\eta_t^2}{2}\right)\|\nabla F(w_{t-1}^S)\|^2\\
		\leq&F(w_{t-1}^S)-\mathbb{E}_{D^S(t)}F(w_{t}^S)+\frac{3\rho\eta_t^2B^2}{2b}+\left(\frac{3\rho\eta_t^2}{2}+\frac{\eta_t}{2}\right)\cdot\frac{2(2+12\sqrt{2})^2B^2 t}{n(T-t)}
		\end{align}}We set $\eta$ to be a constant, and sum the above inequality from $s=1$ to $S$, and $t=1$ to $T$, then we get
	{\small\begin{align*}
		&(\frac{1}{2}\eta-\frac{3\rho\eta^2}{2})\cdot\frac{1}{TS}\sum_{s=1}^S\sum_{t=1}^T\|\nabla F(w_t^s)\|^2\\
		&\leq\frac{F(w_0^1)-F(w^*)}{ST}+\frac{3\rho\eta^2B^2}{2b}+\left(\frac{3\rho\eta^2}{2}+\frac{\eta}{2}\right)\times\left(\frac{1}{T}\sum_{t=1}^T\frac{2(2+12\sqrt{2})^2B^2 t}{n(T-t)}\right).
		\end{align*}}
	By simple calculation, we have  
	{\small$\frac{1}{nT}\sum_{t=1}^{T}\left(\frac{t}{T-t}\right)\leq\frac{\log{T}}{bT}.$}
	
	Let $\eta\leq\frac{1}{6\rho}$, we have $\frac{1}{2}\eta-\frac{3\rho\eta^2}{2}\geq\frac{\eta}{4}$. Then we have
	{\small\begin{align}
		&\frac{1}{TS}\left(\sum_{s=1}^S\sum_{t=1}^T\|\nabla F(w_t^s)\|^2\|^2\right)\\
		\leq&\frac{4(F(w_0^1)-F(w^*))}{\eta(ST+t)}+\frac{6\rho\eta B^2}{b}+2(2+12\sqrt{2})^2B^2\left(6\rho\eta+2\right)\cdot\frac{\log{T}}{bT}\\
		=&\frac{4(F(w_0^1)-F(w^*))}{\eta(ST+t)}+\frac{6\rho\eta B^2}{b}\left(1+\frac{584\log{T}}{T}\right)+1168B^2\cdot\frac{\log{T}}{bT}
		\end{align}}
	
	Let $\eta=\min\left\{\frac{1}{\sqrt{ST}}\cdot\sqrt{\frac{2(F(w_1)-F(w^*))}{\frac{3\rho B^2}{b}\left(1+\frac{584\log{T}}{T}\right)}},\frac{1}{6\rho}\right\}$, we have
	{\small\begin{align}
		&\frac{1}{TS}\left(\sum_{s=1}^S\sum_{t=1}^T\|\nabla F(w_t^s)\|^2\right)\\
		\leq&2\sqrt{\frac{6(F(w_0^1)-F(w^*))\rho B^2}{b(ST+t)}\cdot\left(1+\frac{584\log{T}}{T}\right) }+1168B^2\cdot\frac{\log{T}}{bT}\\
		\leq&\mathcal{O}\left(\sqrt{\frac{(F(w_0^1)-F(w^*))\rho}{bTS}}+\frac{\log{T}}{bT}\right).
		\end{align}}
	By replacing $b$ with $bM$, and using $n=bMT$, we can get the results in the theorem.
	$\Box$
	
	Based on the above theorem, we have the following corollary.
	\begin{corollary}
		If {\small$S<n$}, the convergence rate of distributed SGD with global shuffling {\small$RS_g([n],S,M)$} is comparable with distributed SGD with i.i.d sampling.
		If {\small$S<n$}, distributed SGD with global shuffling {\small$RS_g([n],S,M)$} achieves linear speedup compared with the sequential SGD with {\small$RS_g([n],S,1)$}.
	\end{corollary}
	
	\textit{Proof:}
	
	According to \cite{ghadimi2013stochastic}, the convergence rate for distributed SGD with with-replacement sampling in the nonconvex case is {\small$\mathcal{O}\left(\sqrt{\frac{1}{Sn}}\right)$}. When the first term is dominant, i.e., $S<n$ (the epoch number is smaller than the data size, which is very likely to hold in practice), its order is the same with the convergence rate for global shuffling. 
	
	When $S<n$, the convergence rate for global shuffling is in the order {\small$\frac{1}{\sqrt{Sn}}$}. That is, distributed SGD with global shuffling can achieve linear speedup in the non-convex case.
	
	\subsection{Convergence rate of local shuffling}
	\begin{theorem}
		Conditioned on the partition, the expected convergence rate of distributed SGD with local shuffling in the convex and Lipschitz case is similar to the results given in Theorem \ref{convex thm}, with the term $\sqrt{\frac{1}{n}}$ replaced by $\sqrt{\frac{M}{n}}$; in the strongly convex and smooth case is similar to the results given in Theorem \ref{thm2}, with the term $\frac{\log {n}}{n}$ replaced by $\frac{M\log{n}}{n}$; in the non-convex and smooth case is similar to the results given in Theorem \ref{thm3}, with the term $\frac{\log{n}}{n}$  replaced by $\frac{M\log{n}}{n}$.
	\end{theorem} 
	\textit{Proof sketch:}
	The proof technique for the theorem is similar to that used to prove the results for global random shuffling. Here we just explain their differences.
	
	The first term we need to check is the variance term. For local random shuffling, conditioned on the partition, the variance term becomes
	{\small\begin{align*}
		&\mathbb{E}[\frac{1}{M}\sum^{M}_{m=1}g_{D_m(t)}(w_t)-\frac{1}{M}\sum_{m=1}^M\nabla f_{tb+1:n}^{m}(w_t)]^2\\
		=&\frac{1}{M^2}[\sum_{m=1}^{M}\mathbb{E}[g_{D_m(t)}(w_t)-\nabla f_{tb+1:n}^{m}(w_t)]^2+\sum_{m_1\neq m_2}2cov(g_{D_{m_1}(t)}(w_t),g_{D_{m_2}(t)}(w_t))].
		\end{align*}}Conditioned on the partition, we have $\sum_{m_1\neq m_2}2cov(g_{D_{m_1}(t)}(w_t),g_{D_{m_2}(t)}(w_t))=0$ because they are $i.i.d$ sampled from the underlying distribution $\mathbb{P}$.
	Thus, conditioned on the partition, the variance term can be upper bounded by $\frac{B^2}{Mb}$.
	
	The second term we need to check is the terms which are related to transductive Rademacher Complexity. The term $\mathbb{E}[\frac{1}{M}\sum_{m=1}^Mg_{D_m(t)}(w_t)-\nabla F(w_t)]$ will be changed as follows, if we use local random shuffling instead of global random shuffling,
	{\small$
		\mathbb{E}[\frac{1}{M}\sum_{m=1}^Mg_{D_m(t)}(w_t)-\nabla F(w_t)]=\frac{1}{M}\sum_{m=1}^M\mathbb{E}[g_{D_m(t)}(w_t)-\nabla F^m(w_t)]
		\leq\mathcal{O}(\frac{tb}{n/M}(\frac{1}{\sqrt{tb}}+\frac{1}{\sqrt{n/M-tb}}).$}
	From the above calculation, we can see that the term $\mathbb{E}[\frac{1}{M}\sum_{m=1}^Mg_{D_m(t)}(w_t)-\nabla F(w_t)]$ will be influenced by changing $n$ to $n/M$, which means that it is determined by the local training size $n/M$ instead of the global training size $n$. $\Box$

	\subsection{Insufficient shuffling}
	
	The following lemma generalizes Lemma 3.3 for insufficient shuffling.
	\begin{lemma}\label{lemma6.1}
		Let $\tilde{\sigma}$ be a random permutation over $\{1,\cdots,n\}$ after insufficient shuffling with shuffling error $\epsilon(A,n)\leq\frac{b}{n}$ and $\|s_i\|\leq B$. Then 
		{\small\begin{align}
			\mathbb{E}\left[\frac{1}{n}\sum_{i=1}^ns_i-\frac{1}{bM}\sum_{m=1}^M\sum_{j=1}^{b}s_{\tilde{\sigma}_m(t+j)}\right]\leq\frac{tbM}{n}\cdot\mathbb{E}[s_{1:tbM}-s_{tbM+1:n}]+\frac{4nB\epsilon(A,n)}{n-tMb}.
			\end{align}}
	\end{lemma}
	\textit{Proof:}
	{\small\begin{align*}
		&\mathbb{E}\left[\frac{1}{n}\sum_{i=1}^ns_i-\frac{1}{bM}\sum_{m=1}^M\sum_{j=1}^{b}s_{\tilde{\sigma}_m(t+j)}\right]\\
		=&\mathbb{E}\left[\frac{1}{n}\sum_{i=1}^ns_i-\frac{1}{bM}\sum_{m=1}^M\sum_{j=1}^{b}s_{\sigma_m(t+j)}\right]+\mathbb{E}\left[\frac{1}{bM}\sum_{m=1}^M\sum_{j=1}^{b}s_{\sigma_m(t+j)}-\frac{1}{bM}\sum_{m=1}^M\sum_{j=1}^{b}s_{\tilde{\sigma}_m(t+j)}\right]\\
		\leq&\frac{tbM}{n}\cdot\mathbb{E}[s_{1:tbM}-s_{tbM+1:n}]+B|\mathbb{P}_{v|t}(B_m,m\in[M])-\mathbb{P}_{u|t}(B_m,m\in[M])|\\
		\leq&\frac{tbM}{n}\cdot\mathbb{E}[s_{1:tbM}-s_{tbM+1:n}]+\frac{4nB\epsilon}{n-tbM},
		\end{align*}}
	where the last inequality is using Proposition 2.2.
	
	\begin{theorem}
		Suppose Assumption 1 holds. Distributed SGD with insufficient global shuffling $R_g([n];S,M)$ and learning rate in convex and $L$-Lipschitz continuous case has the following convergence rate,
		{\small$
			\mathbb{E}F(\bar{w}^S)-F(w^*)\leq\mathcal{O}\left\{\frac{1}{\sqrt{Sn}}+\frac{bM}{Sn}+\sqrt{\frac{1}{n}}+\epsilon(A,n)\ln{n}\right\}.
			$}	Set the learning rate $\eta_t^s=\frac{2}{\mu((s-1)T+t)}$ in strongly convex and smooth case, it has the following convergence rate,
		{\small$
			\mathbb{E}\|w^{S}-w^*\|^2\leq\mathcal{O}\left(\frac{\log{n}}{n}+\frac{n\epsilon(A,n)^2}{bM}\right)+\mathcal{O}\left(\min\left\{\frac{bM}{Sn},\frac{\kappa^2(bM)^2\log{Sn}}{(Sn)^2}+\frac{\kappa^2bM\log{n}}{Sn^2}+\frac{\kappa^2n\epsilon(A,n)^2}{SbM}\right\}\right).
			$}
		
		Set {\small$\eta=\min\left\{\frac{1}{\sqrt{ST}}\sqrt{\frac{2(F(w_0^1)-F(w^*))}{\frac{3\rho B^2}{bM}\left(1+\frac{584\log{T}}{T}\right)}},\frac{1}{6\rho}\right\}$} in non-convex case,
		it has the following convergence rate
		{\small$
			\frac{1}{TS}\left(\sum_{s=1}^S\sum_{t=1}^T\|\nabla F(w_t^s)\|^2\right)
			\leq\mathcal{O}\left(\sqrt{\frac{1}{Sn}}+\frac{\log{n}}{n}+\frac{n(\epsilon(A,n))^2}{bM}\right).
			$}
	\end{theorem}
	
	\textit{Proof:} Following the proof of Theorem \ref{convex thm}, Theorem \ref{thm2} and Theorem \ref{thm3}, and using Lemma \ref{lemma6.1} to replace Lemma \ref{lemma3.1}, we can get the results.
	\subsection{With-replacement sampling: strongly convex case}
	\begin{theorem}\label{with}
		Suppose that Assumptions 1 holds. Distributed SGD with with-replacement sampling with mini-batch size $b$, $M$ machines, and $S$ epochs in strongly convex case has convergence rate 
		{\small\begin{equation}
			\mathbb{E}\|w^S-w^*\|^2\leq \mathcal{O}\left(\min\left\{\frac{bM}{nS},\frac{\kappa^2(bM)^2\log{Sn}}{(Sn)^2}\right\}+\frac{1}{Sn}\right)
			\end{equation}}
	\end{theorem}
	\textbf{Proof:}
	At $t+1$-th iteration, we have:
	{\small\begin{align}
		&\mathbb{E}\|w_{t+1}-w^*\|^2\\
		\leq&\|w_t-w^*\|^2-2\mathbb{E}_{b_t}\langle w_{t+1}-w_t,w_t-w^*\rangle+\mathbb{E}_{b_t}\|w_{t+1}-w_t\|^2\\
		=&\|w_t-w^*\|^2-2\eta_t\langle\nabla F(w_t),w_t-w^*\rangle+\eta_t^2\mathbb{E}_{b_t}\|\nabla f_{b_t}(w_t)\|^2\\
		\leq&\|w_t-w^*\|^2-2\eta_t\left(F(w_t)-F(w^*)+\frac{\mu}{2}\|w_t-w^*\|^2\right)\\
		&+\eta_t^2\mathbb{E}_{b_t}\|\nabla f_{b_t}(w_t)-\nabla F(w_t)\|^2+\eta_t^2\|\nabla F(w_t)\|^2\\
		\leq&(1-2\eta_t\mu)\|w_t-w^*\|^2+\eta_t^2\|\nabla F(w_t)\|^2+\frac{\eta_t^2B^2}{bM} \label{ineq6}\\
		\leq&(1-2\eta_t\mu)\|w_t-w^*\|^2+\eta_t^2\left(\frac{B^2}{bM}+G^2\right)
		\end{align}}
	Let $\eta_t=\frac{1}{\mu t}$, we have 
	{\small\begin{align}
		\mathbb{E}\|w_{t+1}-w^*\|^2\leq\left(1-\frac{2}{t}\right)\|w_t-w^*\|^2+\frac{B^2/bM+G^2}{t^2\mu^2}
		\end{align}}
	By induction, we have {\small\begin{equation}\mathbb{E}\|w_{t}-w^*\|^2\leq\frac{B^2/bM+G^2}{t\mu^2}.\label{ineq9}\end{equation}}
	By using the smooth condition, we have 
	{\small\begin{equation}\mathbb{E}\|\nabla F(w_t)\|^2\leq \rho^2\mathbb{E}\|w_{t}-w^*\|^2\leq\frac{\rho^2(B^2/bM+G^2)}{t\mu^2}.\label{ineq10}\end{equation}}
	By putting Ineq.(\ref{ineq10}) into Ineq.(\ref{ineq6}), we have
	{\small\begin{align}
		&\mathbb{E}\|w_{t+1}-w^*\|^2\\
		\leq&\left(1-\frac{2}{t}\right)\|w_t-w^*\|^2+\frac{1}{t^2\mu^2}\cdot\frac{\rho^2(B^2/bM+G^2)}{t\mu^2}+\frac{B^2}{bM\mu^2 t^2}\\
		\leq&\left(1-\frac{2}{t}\right)\|w_t-w^*\|^2+\frac{\rho^2(B^2/bM+G^2)}{t^3\mu^4}+\frac{B^2}{bM\mu^2 t^2}\\
		\leq&\frac{B^2}{bM\mu^2t}+\frac{\kappa^2(B^2/bM+G^2)\log{t}}{t^2\mu^2}
		\end{align}}
	Thus we have
	{\small\begin{align}
		\mathbb{E}\|w_{t+1}-w^*\|^2\leq\mathcal{O}\left(\min\left\{\frac{1}{t},\frac{\kappa^2\log{t}}{t^2}\right\}+\frac{1}{bMt}\right).
		\end{align}}
	Let $t=\frac{Sn}{bM}$, we can get the result in the theorem.
	$\Box$
\end{document}